\documentclass[letterpaper]{article} 
\usepackage{array} 
\usepackage{_style_}  
\usepackage{times}  
\usepackage{helvet}  
\usepackage{courier}  
\usepackage{caption}
\usepackage{newclude}
\usepackage[hyphens]{url}  
\usepackage[colorlinks=true,urlcolor=blue,citecolor=black,linkcolor = black]{hyperref}  
\usepackage[dvipsnames]{xcolor}
\usepackage{graphicx} 
\usepackage{natbib}  
\usepackage{caption} 
\DeclareCaptionStyle{ruled}{labelfont=normalfont,labelsep=colon,strut=off} 
\usepackage{algorithm}
\usepackage{algorithmic}

\usepackage{xspace}

\usepackage{amsmath}
\usepackage{amssymb}
\usepackage{longtable} 
\usepackage{booktabs}

\usepackage{color}
\usepackage{soul}
\usepackage{changes}

\usepackage{multicol}
 
\usepackage{newfloat}

\floatstyle{ruled}
\newfloat{listing}{tb}{lst}{}
\floatname{listing}{Listing}

\author{
    Markus Leippold,\textsuperscript{\rm 1,2}
    Saeid Ashraf Vaghefi,\textsuperscript{\rm 1,3,4}
    Dominik Stammbach,\textsuperscript{\rm 5}
    Veruska Muccione,\textsuperscript{\rm 3, 6}
    Julia Bingler,\textsuperscript{\rm 7, 8}
    Jingwei Ni,\textsuperscript{\rm 1,5}
    Chiara Colesanti-Senni,\textsuperscript{\rm 1}
    Tobias Wekhof,\textsuperscript{\rm 1,10}
    Tobias Schimanski,\textsuperscript{\rm 1}
    Glen Gostlow,\textsuperscript{\rm 1}
    Tingyu Yu,\textsuperscript{\rm 1} 
    Juerg Luterbacher,\textsuperscript{\rm 4}  
    Christian Huggel\textsuperscript{\rm 3}
}
\affiliations{
    \textsuperscript{\rm 1}Department of Banking and Finance, University of Zürich, Zürich, Switzerland\\
    \textsuperscript{\rm 2}Swiss Finance Institute (SFI), Zürich, Switzerland\\
    \textsuperscript{\rm 3}Department of Geography, University of Zürich, Zürich, Switzerland\\
    \textsuperscript{\rm 4}World Meteorological Organization (WMO), Geneva, Switzerland\\
    \textsuperscript{\rm 5}Center for Law and Economics, ETH Zürich, Zürich, Switzerland\\
    \textsuperscript{\rm 6}Swiss Federal Research Institute WSL, Birmensdorf, Switzerland\\
    \textsuperscript{\rm 7}Smith School of Enterprise and the Environment, University of Oxford, Oxford, United Kingdom\\
    \textsuperscript{\rm 8}Council on Economic Policies, Zürich, Switzerland\\
    \textsuperscript{\rm 9}Center of Economic Research ETH Zürich, Zürich, Switzerland\\
 
}

\usepackage{tikz,lipsum,lmodern}
\usepackage[most]{tcolorbox}
\usepackage{color}

\usepackage[utf8]{inputenc}
\usepackage[T1]{fontenc}
\usepackage{setspace}

\usepackage{colortbl} 
\usepackage{longtable}
\usepackage{booktabs}

\definecolor{codegreen}{rgb}{0,0.6,0}
\definecolor{codegray}{rgb}{0.5,0.5,0.5}
\definecolor{codepurple}{rgb}{0.58,0,0.82}
\definecolor{backcolour}{rgb}{0.95,0.95,0.92}

\usepackage{listings,newtxtt}
\newcommand{\judgeClimate}{\textsc{Climinator}}
\newcommand{\medClimate}{\textsc{Climinator}}
\newcommand{\moderator}{{mediator}}
\newcommand{\arbitrator}{{mediator}}
\newcommand{\advocate}{{advocate}}
\newcommand{\advocates}{{advocates}}

\title{Automated Fact-Checking of Climate Change Claims with Large Language Models}

\date{\today} 

\usepackage{verbatim}

\begin{document}

{
\let\pagebreak\relax
\maketitle
}

\begin{abstract}
This paper presents \judgeClimate{}, a novel AI-based tool designed to automate the fact-checking of climate change claims. Utilizing an array of Large Language Models (LLMs) informed by authoritative sources like the IPCC reports and peer-reviewed scientific literature, \judgeClimate{} employs an innovative Mediator-Advocate framework. This design allows \judgeClimate{} to effectively synthesize varying scientific perspectives, leading to robust, evidence-based evaluations. Our model demonstrates remarkable accuracy when testing claims collected from Climate Feedback and Skeptical Science. Notably, when integrating an advocate with a climate science denial perspective in our framework, \judgeClimate{}'s iterative debate process reliably converges towards scientific consensus, underscoring its adeptness at reconciling diverse viewpoints into science-based, factual conclusions. While our research is subject to certain limitations and necessitates careful interpretation, our approach holds significant potential. We hope to stimulate further research and encourage exploring its applicability in other contexts, including political fact-checking and legal domains.
\end{abstract} 

{
\let\pagebreak\relax
\twocolumn
}


\section*{Introduction}
In the ongoing debate on climate change, the truthfulness of public statements is regularly called into question, emphasizing the critical need for swift and reliable fact-checking. A case in point is the recent claim made by Sultan Al Jaber, the president of COP28 and chief executive of the United Arab Emirates’ state oil company Adnoc. On November 21, 2023, Al Jaber controversially asserted that {\it “There is no science out there, or no scenario out there, that says that the phase-out of fossil fuel is what's going to achieve 1.5C.”} This statement, which climate scientists and policymakers immediately challenged, highlights the complexities of separating factual information from opinion in shaping effective, evidence-based climate policy. Recognizing this challenge, 
our paper introduces \judgeClimate{}, a novel framework designed to assess climate-related claims, leveraging advancements in LLMs. \judgeClimate{} -– an acronym for \underline{CLI}mate \underline{M}ediator for \underline{IN}formed \underline{A}nalysis and \underline{T}ransparent \underline{O}bjective \underline{R}easoning -– not only evaluates the accuracy of statements but also enhances its verdicts with evidence-based reasoning and relevant literature references. 


In an era where information proliferates at an unprecedented pace, the task of manually reviewing claims for accuracy becomes increasingly resource-intensive and challenging. Over a decade ago, scholars warned that the exponential growth of online content would eventually overwhelm journalistic fact-checkers, diminishing news quality and contributing to societal harms like diminished government accountability \cite{computational_journalism}. This concern has given rise to a new strand of research in Natural Language Processing (NLP), namely automated fact-checking \citep{computational_journalism, vlachos_riedel, claimbuster, graves_2018, guo-etal-2022-survey}. With misinformation spreading faster and deeper than factual news \cite{vosughi}, there is a pressing need for sophisticated tools capable of effective and real-time fact-checking.

While early automated fact-checking tools, such as those based on the FEVER dataset \cite{thorne2018fever} and climate-focused datasets like climateFEVER \cite{leippold2020climatefever}, have made significant progress, they often fall short in providing the nuanced reasoning necessary for a comprehensive understanding of complex claims. This is where generative AI models, specifically LLMs, come into play. LLMs can offer holistic evaluations rooted in an extensive scientific knowledge base. They can provide the necessary context, reasoning, and argumentation essential for reaching well-informed verdicts on complex climate-related claims. Building on this premise, \judgeClimate{} leverages the capabilities of LLMs to evaluate climate-related claims based on empirical evidence and scientific consensus. The aim is to provide a comprehensive, transparent, and objective assessment of claims that is not limited to countering polarized views but rather enables a more nuanced understanding of climate issues.

\section*{The Pipeline}

\begin{figure*}[h]
\centering
\includegraphics[width=\linewidth]{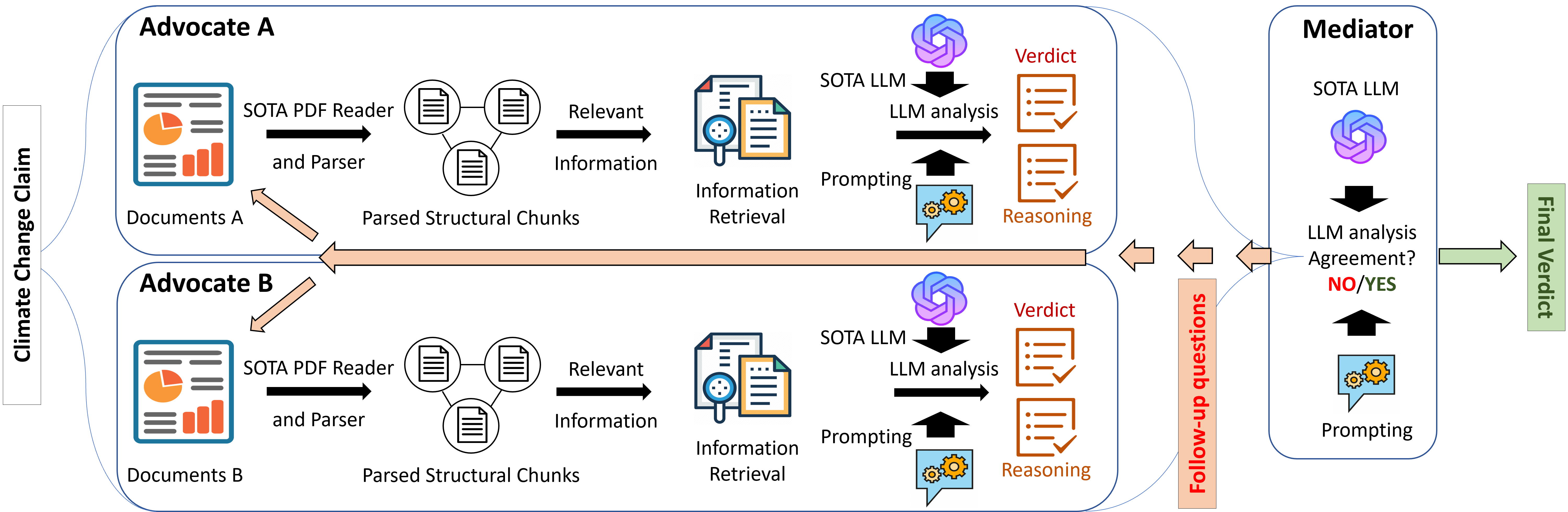}
  \captionsetup{font=small}
    \caption{\medClimate: An LLM-based framework within a Mediator-Advocate system to assess the veracity of climate-related claims.}
\label{fig:pipeline}
\end{figure*}

Figure \ref{fig:pipeline} depicts the operational flow of the \judgeClimate{} framework, where the initial claim is parsed by an LLM into subclaims, enhancing the specificity and efficiency of the evaluation process. Specialized LLMs, henceforth referred to as \advocates, are pivotal in the next phase. Each \advocate{}  examines the claim against a curated corpus of texts. The general GPT-4 model \citep{openai2023GPT4}  serves as one Advocate, while other \advocates{} consist of retrieval-augmented generation (RAG) systems. These systems, which ensure LLM responses are grounded in credible sources, draw from diverse scientific and trusted repositories: the Intergovernmental Panel on Climate Change (IPCC) AR6 reports, World Meteorological Organization (WMO) sources, and two different collections of climate scientists' articles, AbsCC and S1000. We describe the data in Section \ref{sec:data}. Each of the \advocates{} is grounded on one particular text corpus and delivers a verdict informed by its respective data sources, prompted to provide evidence-backed rationales. 

The verdicts, reflecting the different perspectives inherent to each Advocate's text corpus, are consolidated by the Mediator LLM. When \advocates{} concur, the \moderator{} synthesizes these findings into a final judgment. Conversely, disagreements between the \advocates{}  trigger a dynamic dialog that is moderated by the \moderator, who encourages a reassessment and clarification of the positions by asking questions. This iterative exchange persists until a unified verdict is achieved. 
In the design of \judgeClimate{}, we have crafted prompts to evaluate a wide array of climate-related claims, enhancing the tool's versatility and reliability in AI-driven fact-checking. These prompts are thoroughly documented in the appendix, 
ensuring transparency and maintaining the integrity of our approach.

To rigorously test the effectiveness of our \judgeClimate{} framework, we also run experiments in which we introduce an adversarial advocate within the system. This advocate operates under a climate denial stance, processing and challenging information from a contrarian viewpoint. This inclusion serves not to endorse denialism, but to critically examine the robustness of our tool against diverse arguments, including those lacking scientific backing. It aims to identify any potential weaknesses in both the moderated consensus and the scientific literature underpinning it.

\section*{Results}

We apply our approach to different sources of climate-change claims. In particular, we analyze 170 claims fact-checked by \href{https://climatefeedback.org/claimreview/more-than-75-percent-amazon-rainforest-losing-resilience-the-washington-post-sarah-kaplan/}{Climate Feedback}, a website on which a network of climate scientists manually evaluates the credibility of media climate narratives. In the second step, we analyze 163 claims, which are typically made by climate denialists, from the webpage of \href{https://skepticalscience.com/shorturls.php}{Skeptical Science}. 
Finally, we analyze 81 claims from the \href{https://climatechangereconsidered.org/}{Heartland Institute}, which we obtain by extracting the claims from the executive summaries from publications of the so-called NIPCC (Non-governmental International Panel on Climate Change).\footnote{In particular, we use the documents `Climate Change Reconsidered II: Fossil Fuels' and `Climate Change Reconsidered II: Physical Sciences', released on April 9, 2014.} Hence, in total, we analyze a set of 414 claims from different sources and of different complexity.

\subsection*{Climate Feedback}

Climate Feedback categorizes the verdicts into twelve different categories, ranging from \texttt{incorrect} to \texttt{correct}, with different shades of correctness and accuracy in between. A typical example from Climate Feedback is, e.g., the following claim:

\begin{tcolorbox}[colback=blue!15!white,colframe=blue!15!white]\scriptsize \sffamily 
More than 75 percent of the rainforest is losing resilience. More than half of the rainforest could be converted into savanna in a matter of decades. \\
  \rightline{\textit{\underline{source}}: Washington Post, March 15, 2022}\medskip \\ 
    {\textit{Climate Feedback Verdict}: \color{OliveGreen}{mostly accurate}} 
\end{tcolorbox}

This claim obtained the verdict ``mostly accurate'' on \href{https://climatefeedback.org/claimreview/more-than-75-percent-amazon-rainforest-losing-resilience-the-washington-post-sarah-kaplan/}{Climate Feedback}. A climate scientist usually provides the verdict and the reasoning, writing a few paragraphs with references to the relevant literature. In this case, the scientist argues that the claim is accurate in the sense that recent scientific findings indicate that a substantial portion, over 75 percent, of the Amazon rainforest has indeed experienced a decrease in resilience since the early 2000s. However, they note that the term "savanna-like" is employed to describe the potential transition of the Amazon rainforest rather than a complete transformation into a savanna ecosystem. The key distinction lies in the reduced diversity of tree species and a diminished capacity to store carbon in the altered Amazon ecosystem as compared to a typical savanna ecosystem. This verdict exemplifies the subtleties involved in providing a verdict to a given claim. 


\begin{figure}[h]
\centering
\includegraphics[width=1\linewidth]{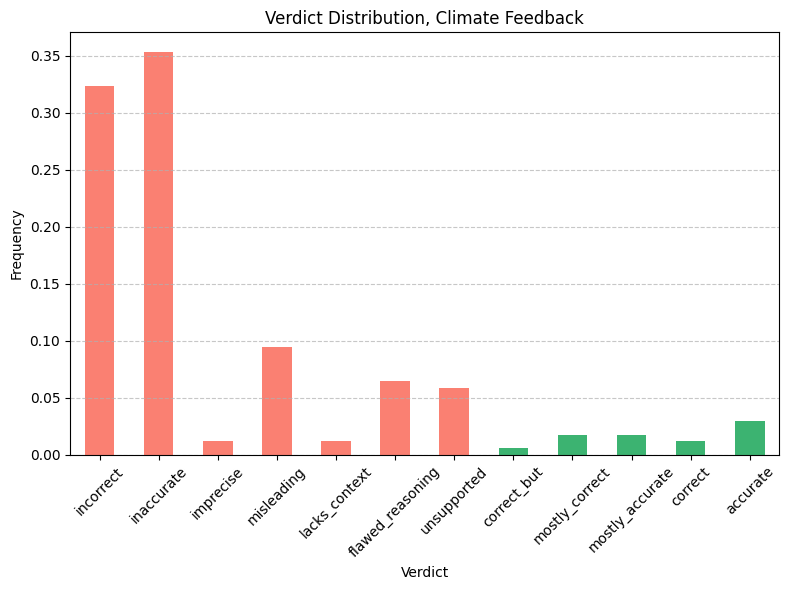}
  \captionsetup{font=small}
    \caption{Distribution of claim verdicts for Climate Feedback, annotated by climate experts. The red bars indicate the verdicts that end up as ``incorrect'', while the green bars end up as ``correct''. \label{fig:cf_verdicts}}
\end{figure}

\begin{figure}[t]
\centering
\includegraphics[width=0.9\linewidth]{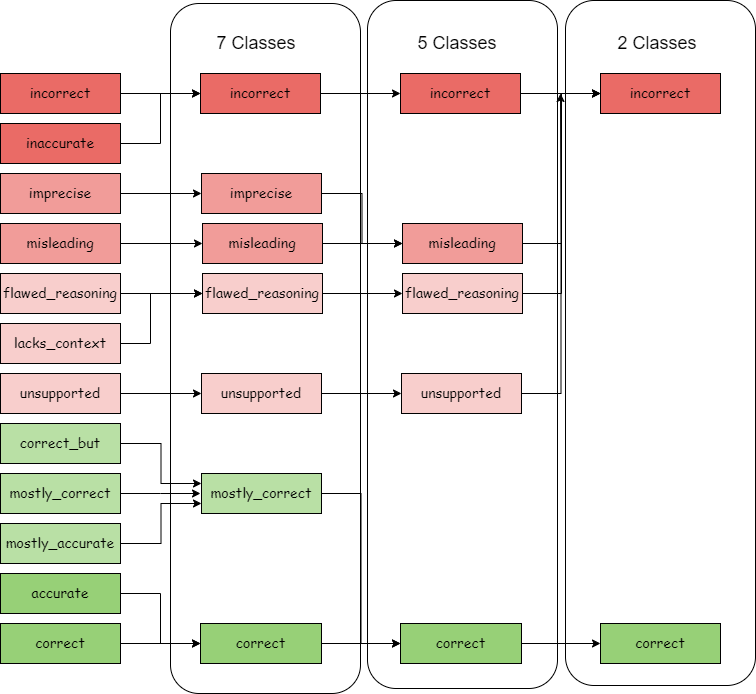}
  \captionsetup{font=small}
    \caption{Three levels of category consolidation, starting from the original twelve Climate Feedback verdicts.\label{fig:cons}}
\end{figure}

\paragraph{Consolidating the verdicts.} Figure \ref{fig:cf_verdicts} gives an overview of the verdicts of the different models. Climate Feedback provides twelve distinct verdict categories, which might generate unnecessary complexity. Therefore, to streamline the process, we consolidated these fine-grained categories into broader classes. The consolidation process consists of three successive stages: In the initial stage, we reduced the categories to seven. In the second stage, we further condensed them to five, and finally, we merged them into a binary classification comprising only 'correct' and 'incorrect' categories (see Figure \ref{fig:cons}). This stratification into different levels of categorization complexity helps in elucidating the nuanced capabilities of the models in handling tasks ranging from highly detailed to broadly binary classifications.\footnote{We add the additional category \textsc{NEI} (not enough information) when the LLMs think there's insufficient information to come up with a verdict. However, for the performance analysis of the different models, we discard the \textsc{NEI} class for better comparison.}

\begin{figure}[ht]
\centering
\includegraphics[width=\linewidth]{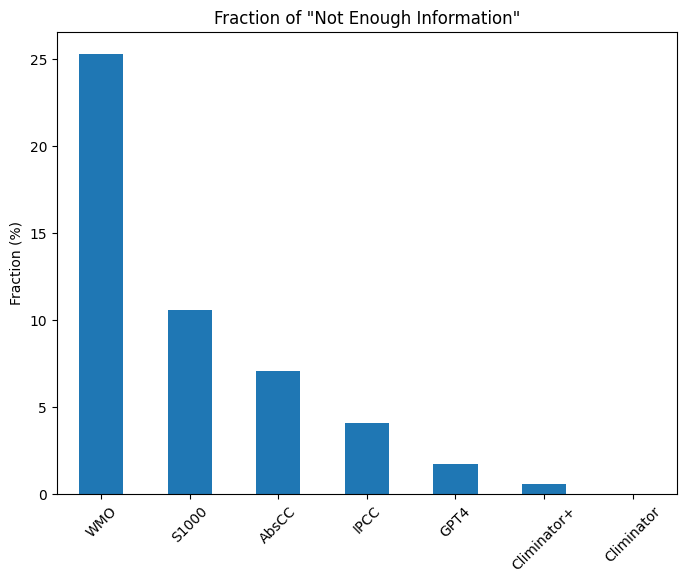}
  \captionsetup{font=small}
    \caption{ The ratio of `Not Enough Information' (\textsc{NEI}) for different models. The model generates a \textsc{NEI}, if it cannot access the information needed to provide a verdict on the claim taken from Climate Feedback.}
\label{fig:neis_CF}
\end{figure}

Before we analyze the results in detail, we show the fraction of cases in which a given model fails to come to a verdict in Figure \ref{fig:neis_CF}. Such cases occur when the LLM cannot find enough evidence in the retrieved text to support or refute a verdict. The \advocate{} of the WMO has a high fraction of such \textsc{NEI}'s (25.3\%). This does not come as a surprise, as its text corpus is relatively small compared to the other \advocates.

\paragraph{Performance analysis.} Table \ref{tab:table1} presents an analysis of the performance of various models in classifying Climate Feedback claims. By \judgeClimate, we denote the Moderator-Advocate system that includes all the RAG Advocates (IPCC, WMO, AbsCC, and S1000). \judgeClimate{+} includes all the \advocates above but also includes the plain GPT-4 model without any retrieval as an additional advocate. The models have been evaluated on their ability to classify claims as correct or incorrect using varying levels of granularity, as structured within the six panels of the table. The classification is based on the different consolidated categories outlined in Figure \ref{fig:cons}, with Panel A consisting of seven classes, Panel B five classes, and Panel C two classes. The performance metrics reported include Precision (Prec.), Recall (Rec.), macro-F1 values (F1), and Accuracy (Acc.), all widely used and standard indicators of model effectiveness.


Panels A.1, B.1, and C.1 of Table \ref{tab:table1} exclude the 'Not Enough Information' (\textsc{NEI}) category. In these panels, we show that \judgeClimate{} and \judgeClimate{+} are the most effective at automatically fact-checking climate change claims. We find that GPT-4, without any further climate-related information support, performs worst in the case of 8 and 5 classes -- all individual advocates are more effective at fact-checking. Specifically, \judgeClimate{}, through its mediating process, leads with an accuracy of 62.94\%, 64.71\%, and 97.06\% across the three respective panels, indicating a robust capability in discerning among the multi-class scenarios. The IPCC model shows competitive accuracy, particularly in Panels A.1 and B.1. GPT-4 without retrieval-augmented information by far achieves the lowest Recall rates. This suggests a cautious approach, favoring the exclusion of false positives errors at the expense of encountering more false negatives errors in the classification process. However, this cautious approach results in the lowest overall accuracy for the 8 and 5 classes.

When incorporating the \textsc{NEI} category in Panels A.2, B.2, and C.2 of Table \ref{tab:table1}, the data presents a nuanced view of model adaptability to claims of uncertain veracity. In these panels, both \judgeClimate{} and \judgeClimate{+} maintain commendable Precision, Recall, F1, and Accuracy scores, reinforcing their robustness in discerning accurate categorizations. However, the inclusion of NEI introduces a discernible challenge for the IPCC and WMO models, with a notable drop in accuracy. GPT-4’s performance further declines in these panels, evidenced by reduced Precision and Recall.


\begin{table*}[h!]
  \begin{center}
  \captionsetup{font=small}
    \caption{ Performance metrics for 170 Climate Feedback claims. The table reports Precision (Prec.), Recall (Rec.), macro-F1 values (F1), and Accuracy (Acc.) for the different models relative to the verdicts provided by the scientists at Climate Feedback. From the original 13 categories of verdicts, we merge the categories into seven (Panel A), five (Panel B), and two (Panel C) classes, according to Figure \ref{fig:cons}.}
    \label{tab:table1}
      \scalebox{0.85}{
    \begin{tabular}{lrrrrrrrrrrrr} \toprule
       & \multicolumn{4}{c}{{Panel A.1: 7 classes  excl. \textsc{NEI}}} & \multicolumn{4}{c}{{Panel B.1: 5 classes excl. \textsc{NEI} }} & \multicolumn{4}{c}{{Panel C.1: 2 classes excl. \textsc{NEI}}} \\
     \cmidrule(lr{.75em}){2-5}  \cmidrule(lr{.75em}){6-9} \cmidrule(lr{.75em}){10-13}
     {Model} & {Prec.} & {Rec.} & {F1} & {Acc.} & {Prec.} & {Rec.} & {F1} & {Acc.} &  {Prec.} & {Rec.} & {F1} & {Acc.}\\
     \cmidrule(lr{.75em}){1-5}  \cmidrule(lr{.75em}){6-9} \cmidrule(lr{.75em}){10-13}
    \judgeClimate{} (170)  & 42.74 & \textbf{41.19} & \textbf{38.87} & \textbf{62.94}  & 44.35 & \textbf{46.14} & \textbf{43.84} & \textbf{64.71} &  89.35 & \textbf{91.90} & \textbf{90.58} & \textbf{97.06}      \\
    \judgeClimate{+} (169) & 40.66 & 35.05 & 34.71 & 62.72  & 41.66 & 40.40 & 40.60 & 64.50 &  \textbf{94.19} & 85.39 & {89.20} & {97.04}   \\ \midrule
    IPCC  (163)            & 34.13 & 37.39 & 34.48 & {61.96}  & 36.71 & {42.04} & 38.85 & {63.80} & 83.69 & {90.64} & {86.75} & 95.71     \\
    WMO (127)              & 35.11 & 29.53 & 24.98 & 54.33  & \textbf{50.75} & 41.08 & 36.17 & 56.69 & 66.89 & 75.65 & 70.10 & 92.13     \\
    AbsCC (158)            & 26.41 & 29.64 & 25.71 & 50.63  & 34.75 & 42.00 & 36.08 & 56.96 & 72.32 & 88.00 & 77.10 & 89.87      \\  
    S1000 (152)            & 32.26 & 36.77 & 32.59 & 56.58  & 36.89 & 43.07 & 38.20 & 58.55 & 73.88 & 84.29 & 77.81 & 92.11           \\ \midrule
    GPT-4 (167)            & \textbf{43.48} & 27.66 & 28.36 & 49.70  & 45.84 & 33.54 & 35.52 & 52.10 & 88.41 & 80.12 & 83.66 & 95.81         \\ \midrule

       & \multicolumn{4}{c}{{Panel A.2: 7 classes incl. \textsc{NEI}}} & \multicolumn{4}{c}{{Panel B.2: 5 classes  incl. \textsc{NEI}}} & \multicolumn{4}{c}{{Panel C.2: 2 classes  incl. \textsc{NEI}}} \\
     \cmidrule(lr{.75em}){2-5}  \cmidrule(lr{.75em}){6-9} \cmidrule(lr{.75em}){10-13}
     {Model} & {Prec.} & {Rec.} & {F1} & {Acc.} & {Prec.} & {Rec.} & {F1} & {Acc.} &  {Prec.} & {Rec.} & {F1} & {Acc.}\\
     \cmidrule(lr{.75em}){1-5}  \cmidrule(lr{.75em}){6-9} \cmidrule(lr{.75em}){10-13}
    \judgeClimate{}    & \textbf{42.74} & \textbf{41.19} & \textbf{38.87} & \textbf{62.94}  & \textbf{44.35} & \textbf{46.14} & \textbf{44.27} & \textbf{64.71} & 89.35 & \textbf{91.90} & \textbf{90.58} & \textbf{97.06}  \\
    \judgeClimate{+}  & 40.66 & 34.95 & 34.66 & 62.35  & 41.66 & 40.26 & 40.53 & 64.12  & \textbf{94.19} & 85.07 & {89.04} & 96.47  \\ \midrule
    IPCC       & 29.86 & 31.04 & 29.10 & 59.41  & 30.59 & 32.99 & 31.44 & 61.18  & 55.80 & 57.17 & 56.35 & 91.76        \\
    WMO        & 27.31 & 17.75 & 17.06 & 40.59  & 42.30 & 24.38 & 25.23 & 42.35  & 44.59 & 33.67 & 38.36 & 68.82  \\
    AbsCC      & 20.54 & 21.85 & 19.44 & 47.06  & 28.96 & 33.18 & 29.22 & 52.94  & 48.21 & 56.35 & 50.09 & 83.53   \\  
    S1000      & 25.09 & 24.78 & 23.57 & 50.59  & 30.74 & 31.18 & 29.66 & 52.35  & 49.25 & 49.42 & 48.86 & 82.35   \\ \midrule
    GPT-4      & 33.82 & 21.13 & 21.58 & 48.82  & 38.20 & 27.04 & 29.00 & 51.18  & 58.94 & 51.53 & 54.60 & 94.12  \\ \bottomrule
    \end{tabular}}
  \end{center}
\end{table*}

In summary, the analysis reveals that \judgeClimate{} and \judgeClimate{+} are strong contenders, excelling in nuanced categorization tasks while maintaining high accuracy. IPCC and WMO demonstrate competitive performance, particularly in multi-class scenarios, albeit with some trade-offs. AbsCC and S1000 are effective in less granular tasks. Clearly, the choice of class configuration impacts model performance, highlighting the trade-offs between granularity and accuracy in climate claim categorization. Higher granularity in categorizing claims can lead to more detailed and specific insights. However, this inevitably reduces overall accuracy due to the complexity of making fine distinctions between multiple categories.

A high granularity may be crucial in contexts where nuanced understanding is essential, such as in academic research, policy formulation, or advanced educational settings. In these scenarios, detailed categorization can shed light on the specific nature of inaccuracies or misconceptions, aiding in targeted corrective measures or in-depth analysis. However, for broader public communication or preliminary screenings, where the primary goal is to quickly identify and address blatantly false claims, such high levels of detail may not be necessary.

\paragraph{Analyzing diverging verdicts.} Fact-checking the 170 claims on Climate Feedback, all the models reached a final verdict after the first round. Hence, there was no debating taking place. We argue that the absence of debating is due to the fact that the claims are clearly defined and, given that all the advocates draw their information from text sources that are grounded on climate science, have inherently low disagreement. Therefore, the \moderator{} does not ask for an additional debating round.

Although there is no debating, it may still be instructive to analyze the few cases in which \judgeClimate{} and the verdict from the scientists from Climate Feedback diverge when we analyze the claims using the binary ``correct'' versus ``incorrect'' classification. As an illustrative example, we take the following claim:

\begin{tcolorbox}[colback=blue!15!white,colframe=blue!15!white]\scriptsize \sffamily 
Amazon near tipping point of switching from rainforest to savannah. \\
  \rightline{\textit{\underline{source}}: Fiona Harvey, The Guardian, 5 Oct. 2020}\medskip \\ 
    {\textit{Climate Feedback Verdict}: \color{red}{imprecise}} \\
     {\judgeClimate: \color{OliveGreen}{mostly accurate}}    
    \end{tcolorbox}

To reach a verdict, \judgeClimate{} first notes that while the IPCC, AbsCC, and S1000 LLMs offer substantial backing, the WMO concurs only on the drying trend, citing a lack of information on the tipping point. The primary caveat in the final assessment is the uncertainty regarding the precise timing of this ecological shift, although the overall direction towards a tipping point is clear. Hence, \judgeClimate{} reaches the verdict that the claim is mostly accurate, with the noted uncertainty about the timing. This verdict, according to the aggregation in Figure \ref{fig:cons}, merges into the category ``correct''. 

In contrast, Climate Feedback marks the claim as ``imprecise'' due to a lack of context and potential misinterpretation of the findings in \citet{staal2020hysteresis}, to which the quoted claim refers. The cited study does not support the immediacy suggested by the claim but rather discusses forest resilience and the existing climatic potential for savanna conditions. The complexity of forest-rainfall feedback mechanisms and the gradual nature of ecological transitions are not adequately conveyed, leading to an imprecise portrayal of the Amazon's current state. Thus, the verdict is ``imprecise,'' which then collapses to ``incorrect'' in our setting (see Figure \ref{fig:cons}). 

The evaluation of the claim regarding the Amazon's potential tipping point from rainforest to savannah reveals a divergence in focus between Climate Feedback and \judgeClimate{}. While \judgeClimate{} addresses the claim as quoted above, which can be found on the overview page of Climate Feedback, Climate Feedback examines a more expanded version of the claim in its discussion, including details about the percentage of the rainforest likely to turn into savannah and the processes involved in the shift.\footnote{For a comparison, see the link with the short claim \href{https://climatefeedback.org/claim-reviews/6/}{here} (which is used for \judgeClimate), and compare it with the longer claim \href{https://climatefeedback.org/claimreview/climate-change-can-make-it-harder-for-the-amazon-rainforest-to-grow-back-from-deforestation-but-that-does-not-mean-40-of-it-will-now-turn-into-a-savanna/}{here}.} The verdict `imprecise' primarily addresses the percentage of the rainforest turning into savannah and hence suggests an oversimplification. The verdict thus emphasizes a lack of necessary context, particularly as it implies an immediacy not fully supported by the source study. Conversely, the "mostly accurate" verdict, applied to the broader assertion of the Amazon nearing a tipping point, acknowledges substantial scientific support for this phenomenon. In this context, \judgeClimate could provide insights into the likelihood and occurrence of a tipping point, which is a high relevant question in the development of nature-realted scenarios. In particular, by assessing the accuraccy of statements related to this topic, the algorithm could help assess the likelihood of a tipping points under different beliefs about the future. This would be of practical relevance in that users could be able to pick a scenario (with a given likelihood of occurrence) depending on their beliefs about fure developments (e.g. more or less optimistic).

From the above discussion, it becomes clear that decomposing the claim into its constituent parts — the likelihood of a tipping point and the specific percentage of the forest affected — allows for a more detailed and accurate evaluation of its various elements. Hence, we feed in the whole claim as eventually analyzed by Climate Feedback into \judgeClimate{} and ask the LLM model to first decompose the claim into its part. The aggregated verdict is still ``mostly accurate'' but with the nuanced comment:

\begin{tcolorbox}[colback=orange!15!white,colframe=orange!15!white]\scriptsize
The summary of the verdict is that the claim is mostly accurate, with the caveat that the specific figure of 40\% is not fully supported by the evidence provided. The sensitivity of rainforests to changes in rainfall and moisture, the impact of fires and droughts, and the gradual nature of the transition from rainforest to savannah are well-supported by the evidence. The difficulty in reversing the process once it has begun is also supported, although specific details on the timescale and the exact percentage of the Amazon at risk are less certain. Therefore, the overall claim is ([[mostly\_accurate]]), with a high level of evidence and agreement for the sensitivity to rainfall and moisture, and the impact of fires and droughts, but a lower level of confidence in the specific figure of 40\% and the timescale of the transition.
\end{tcolorbox}

Interestingly, the lead author of the study referenced in the claim noted on Climate Feedback that the statement “in itself is not a false representation of our work,” suggesting that while specific aspects of the claim may lack precision, the overall concept is not inaccurately portrayed. Hence, the accuracy of \judgeClimate{} would arguably be even higher than reported in Table \ref{tab:table1}. In Appendix \ref{sec:amazon}, we present all the output generated by the \judgeClimate{} pipeline for the above claim. We also list the additional claims that led to a discrepancy between Climate Feedack and \judgeClimate.


\subsection*{Skeptical Science}

Skeptical Science is one of the first and arguably the most popular platforms for verifying climate change claims. The typical claims on this platform are often of a more colloquial nature than on Climate Feedback. For example, Skeptical Science analyzes claims such as:

\begin{tcolorbox}[colback=blue!15!white,colframe=blue!15!white]\scriptsize \sffamily 
   We are heading into an ice age.\\
  \rightline{\textit{\underline{source}}: Skeptical Science}
\end{tcolorbox} 

Such vague claims leave a lot of ambiguity and room for interpretation and may, therefore, pose problems for automated claim verification models to reach a final verdict.

\begin{figure}[t]
\centering
\includegraphics[width=1\linewidth]{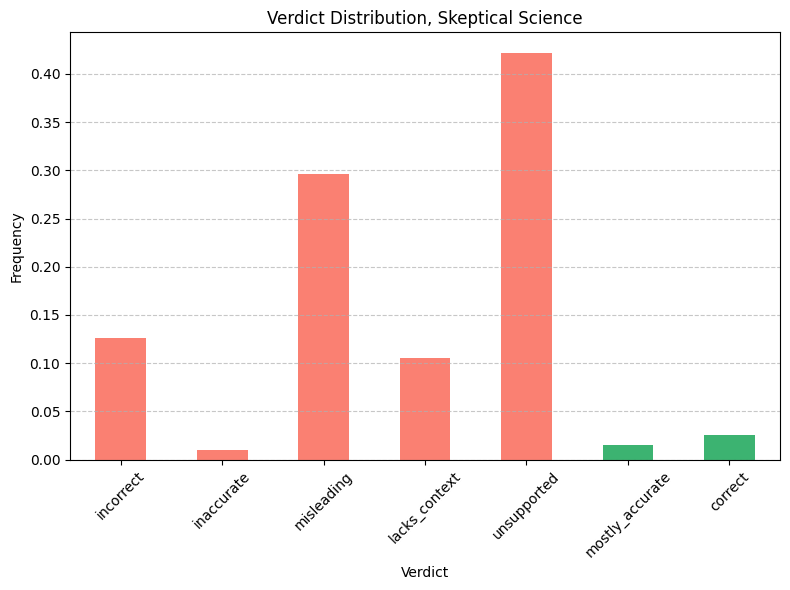}
  \captionsetup{font=small}
    \caption{ Distribution of claim verdicts for Skeptical Science. The red bars indicate the verdicts that end up as ``incorrect'', while the green bars end up as ``correct'' (see Figure \ref{fig:cons}). \label{fig:sks_verdicts}}
\end{figure}

\paragraph{Obtaining claim verdicts.} Compared to Climate Feedback, Skeptical Science does not provide an annotated label for the verdict. To obtain a label, we therefore proceed as follows. We feed the claims with the scientific explanations from Skeptical Science and ask GPT-4 to evaluate the claim in relation to the explanations given, without using any further information. To simplify compatibility, we use the same set of possible verdicts as for Climate Feedback (see figure \ref{fig:cons}), and we plot their distribution in Figure \ref{fig:sks_verdicts}. Based on this classification, we then simplify the dataset to a binary classification task, following the procedure laid out in Figure \ref{fig:cons}.

\begin{figure}[ht]
\centering
\includegraphics[width=\linewidth]{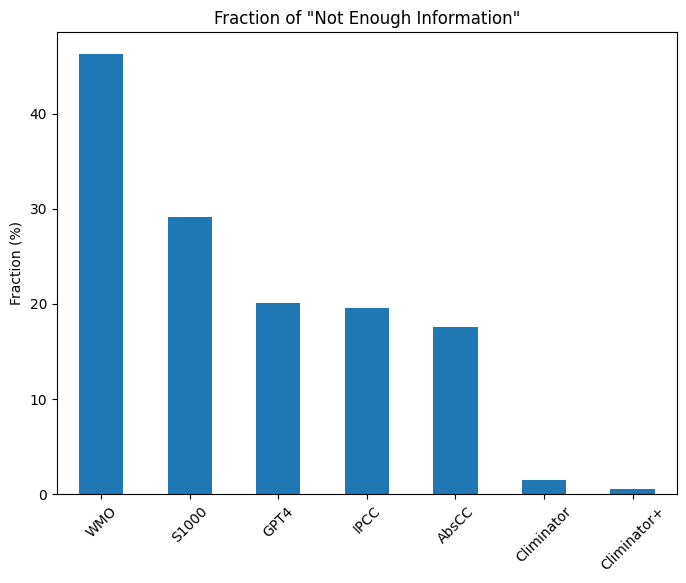}
  \captionsetup{font=small}
    \caption{The ratio of `Not Enough Information' (\textsc{NEI}) for different models. The model generates a \textsc{NEI}, if it cannot access the information needed to provide a verdict on the claim taken from Skeptical Science.}
\label{fig:neis_Sk}
\end{figure}

\paragraph{Explaining the fraction of NEIs.} The vagueness of the claims on Skeptical Science may cause some problems for automated fact-checking systems. Figure \ref{fig:neis_Sk} shows the relative fraction of cases in which a given model cannot come to a verdict due to a lack of sufficient information about a given claim. We observe that, for Skeptical Science claims, the WMO \advocate{} struggles even more than for Climate Feedback claims to reach a conclusion, with the highest \textsc{NEI} fraction at 46.23\%. Surprisingly, even GPT-4 has a very high rate of \textsc{NEI}'s with 20.1\%, compared to \medClimate{} and \medClimate{+} with only 1.5\% and 0.5\%, respectively. Such a more assertive stance in categorization could potentially lead to an increased misclassification risk, highlighting a fundamental trade-off between the models' decisiveness and risk aversion in classification tasks.

A straightforward question to ask next is: Why do we observe such differences in the reported amount of \textsc{NEI}'s in the two datasets, the claims from Climate Feedback and Skeptical Science. In Appendix \ref{app:complex}, we analyze the semantic and syntactic differences of the claims from the different sources available. We find that the higher incidence of \textsc{NEI}-classifications for Skeptical Science is due to its broad but shallow approach. The lower syntactic complexity might strip away the contextual or semantic depth needed for a model to make a confident, nuanced assessment. Interestingly, this is particularly the case for GPT-4. To reduce the impact of NEI's, we therefore include the statement ``This claim is made in a climate-change context: [claim]," so that the model can better capture the context of the claim.

\paragraph{Performance analysis.} Table \ref{tab:skeptical} presents the accuracy of various models in the binary classification of climate-change-related claims, using data collected from Skeptical Science. The table reports two sets of accuracy metrics: "Acc. excl. NEI" and "Acc. incl. NEI". 

Notably, \judgeClimate{} achieves the highest accuracy of 99.38\% (excl. NEI) and 98.77\% (incl. NEI) among all models, demonstrating close to perfect classification accuracy for automatically fact-checking climate-change-related claims. \judgeClimate{+} also performs admirably with an accuracy of 96.93\% in both scenarios. The IPCC model achieves an accuracy of 96.99\% (excl. NEI) but experiences a notable drop to 79.14\% when including NEI claims, highlighting a significant impact of NEI categorizations on its performance. Similarly, WMO demonstrates an outstanding accuracy of 98.80\% (excl. NEI) but experiences a substantial decrease to 50.31\% when NEI claims are included. AbsCC and S1000 perform reasonably well, with accuracies above 93\% when excluding NEI, but their performance again decreases when including NEI claims. Lastly, GPT-4 achieves an accuracy of 95.14\% (excl. NEI) and 84.05\% (incl. NEI), showcasing a noticeable decrease when NEI categorizations are considered. These results emphasize the importance of handling NEI instances effectively in such tasks.

\begin{table}[h!]
  \begin{center}
  \captionsetup{font=small}
    \caption{Accuracy for climate-change claims collected from Skeptical Science. The table reports the Accuracy (Acc.) for the different models relative to the verdicts provided by the Skeptical Science. The results are based on binary classification (correct/incorrect) }
    \label{tab:skeptical}
    \scalebox{0.825}{ 
    \begin{tabular}{lrrrr} \toprule
      Model & Acc. ex NEI & \# Claims  &  Acc. & \# Claims  \\ \midrule
      \judgeClimate{} & \textbf{99.38} &  162 &\textbf{98.77}& 163 \\
      \judgeClimate{+} & 96.93 & 163  & 96.93 & 163\\  \midrule
      IPCC & 96.99 & 133& 79.14 & 163\\
      WMO & 98.80 & 83&50.31& 163 \\
      AbsCC & 93.94 & 132&76.07& 163 \\
      S1000 & 96.55 & 116 &68.71 & 163\\  \midrule
      GPT-4 & 95.14 & 144& 84.05& 163 \\ \bottomrule
    \end{tabular}
    }
  \end{center}
\end{table}

Our analysis suggests that \judgeClimate{} diverges only in one of the Skeptical Science claims. The claim is the following:

\begin{tcolorbox}[colback=blue!15!white,colframe=blue!15!white]\scriptsize \sffamily 
Hansen's 1988 prediction was wrong.\\
  \rightline{\textit{\underline{source}}: Skeptical Science}
\end{tcolorbox}


Skeptical Science categorizes this claim as ``incorrect.'' They explain that although Hansen's prediction of a global temperature increase was higher than the actual warming observed, this discrepancy stems from his use of a high climate sensitivity parameter. Had Hansen employed the currently accepted value of approximately 3°C warming for a doubling of atmospheric CO2, his model would have more accurately projected the observed global warming.

In contrast, \judgeClimate{}'s assessment posits:

\begin{tcolorbox}[colback=orange!15!white,colframe=orange!15!white]\scriptsize
[...] The final verdict, synthesizing the information from the Advocates, is that the claim is [[mostly\_accurate]] but lacks context regarding the reasons for the overestimation and the general correctness of the warming trend predicted by Hansen once actual forcings are considered.
\end{tcolorbox}

While \judgeClimate{} concludes with a ``mostly accurate'' verdict, it acknowledges the absence of context concerning Hansen's overestimation and the broader accuracy of his warming trend prediction when actual forcings are taken into account. This perspective, despite differing in classification from Skeptical Science's ``incorrect'' label, is not fundamentally at odds with Skeptical Science's assessment. Both analyses underscore the significance of understanding the underlying assumptions in Hansen’s model, thereby highlighting that \judgeClimate{}'s evaluation, though different in its final verdict, aligns in principle with the critical assessment provided by Skeptical Science.

\paragraph{Absence of follow-up questions.} Interestingly, in none of the claims analyzed so far did the \arbitrator{} ask specific ``follow-up questions.'' An obvious reason for the absence of such a verdict is the overwhelming consensus among climate experts \citep[e.g.,][for a recent survey]{Myers2021ConsensusRQ}. To simulate a debate with climate science denialist opinions, and to foster a more comprehensive understanding of the model performance when being challenged by opposing views, we introduce an adversarial advocate. This advocate would represent the views of climate denial, debating the scientific consensus and thereby challenging the discourse.

\subsection*{Adding the Climate-Denier Perspective}  


\begin{table*}[t]
\centering
  \captionsetup{font=small}
\caption{Verdict frequencies for Climate Feedback claims including adversarial NIPCC advocate.  By \judgeClimate{} (A), we denote the original model that is enriched with an NIPCC advocate. By \judgeClimate{} (B), we denote the model that has only two advocates, the IPCC and the NIPCC.}
\label{tab:matrixcf}
      \scalebox{1}{
\begin{tabular}{lcccccc}
\hline
\toprule 
Verdict & NIPCC &  \judgeClimate& \multicolumn{2}{c}{\judgeClimate{}(A)} & \multicolumn{2}{c}{\judgeClimate{}(B)} \\
  &  &  & Round 1 & Round 2& Round 1 & Round 2  \\
\midrule 
incorrect & 57 & 155 & 136 & 154 & 98 & 156 \\ 
correct & 81 & 15 & 7 & 16 & 6 & 14 \\ 
follow-up & 0 & 0 & 25 & 0 & 64 & 0 \\  \midrule
\textsc{NEI} & 32 & 0 & 2 & 0 & 0 & 0 \\
\bottomrule 
\end{tabular}}
\end{table*}

The NIPCC\footnote{\url{http://climatechangereconsidered.org/}} is an entity associated with the organized climate denial movement. Established by Fred Singer in 2004, the NIPCC positions itself as an alternative to the Intergovernmental Panel on Climate Change (IPCC), spreading false information about the IPCC itself and climate science. The NIPCC's reports, which are not peer-reviewed, are disseminated by the Heartland Institute—a politically partisan organization known for its active promotion of climate change denial. To construct a RAG model, we utilize two NIPCC reports: \cite{idso2013climate}, focusing on physical science, and \cite{idso2019climate}, concentrating on fossil fuels. This RAG model, dubbed the NIPCC \advocate{}, is designed to echo the climate denial perspective. We also remark that for the following experiments, we changed the prompt for the \moderator{} to avoid potential biases in favor of climate science, i.e., we force the model to only mediate between the advocates and keep it from a biased view given its internal knowledge (a detailed description can be found in Appendix \ref{sec:Prompting}).

\paragraph{Analyzing the impact of the NIPCC Advocate.} In our experiment, we re-analyzed Climate Feedback claims by incorporating the NIPCC \advocate{}. Table \ref{tab:matrixcf} displays a heatmap of the claims as predicted by various models. Notably, the NIPCC \advocate{} categorizes 57 claims as incorrect and 81 as correct, a stark contrast to the \judgeClimate{} model, which identifies 155 claims as incorrect and 15 as correct. The inclusion of the NIPCC \advocate{} into the \judgeClimate{} framework, resulting in the \judgeClimate{} (A) variant, significantly alters the outcome: only 136 claims are marked correct, and 25 are moved to the "follow-up question" category. 

Obviously, the contrarian view of the NIPCC advocates triggers a debate process where the \moderator{} poses follow-up questions to the \advocates{}, and their responses, along with initial answers, are reevaluated by each \advocate{}. Subsequently, they are asked to reconsider their initial verdicts. The \moderator{} then reassesses these inputs to finalize the verdict or poses new questions. In this iteration, no further follow-up questions were needed after the second round, leading \judgeClimate{} (A) to classify 154 claims as incorrect.

This outcome might be attributed to the imbalance in the number of advocates representing the scientific consensus versus the lone climate science denier advocate, who aims to spread misinformation. To check this, we introduce \judgeClimate{} (B), a model solely informed by the texts of the IPCC and the NIPCC. In the first round, this model labels only 98 claims as incorrect. However, despite the increased influence of the NIPCC, it does not surpass \judgeClimate{} (A) in terms of the number of correct claims, with a substantial number (64) requiring follow-up questions. Upon progressing to the second round, most of these ambiguous claims are reclassified as incorrect, culminating in a final count of 156 incorrect claims and 14 correct ones. This distribution closely mirrors the original \judgeClimate{} model's results.

\begin{table}[h!]
  \begin{center}
  \captionsetup{font=small}
    \caption{Performance metrics of models on Climate Feedback claims with NIPCC. By \judgeClimate{} (A), we denote the original model that is enriched with an NIPCC advocate. By \judgeClimate{} (B), we denote the model that has only two advocates, the IPCC and the NIPCC. We do not filter out the \textsc{NEI} and `follow-up question' cases.}
    \label{tab:CF_NIPCC}
    \scalebox{0.9}{ 
    \begin{tabular}{lrrrr}\toprule
      {Model} & Prec. & Rec. & F1 & Acc. \\
      \midrule
    \judgeClimate{} & 89.35 & 91.90 & 90.58 & 97.06 \\   \midrule  
      \judgeClimate{}(A) & 54.59 & 81.01 & 50.90 & 83.53 \\
      Round 2 & 83.40 & 87.68 & 85.38 & 95.29\\  \midrule  
      \judgeClimate{}(B) & 36.99 & 70.74 & 26.40 & 58.24\\
      Round 2 & 84.43 & 84.43 & 84.43 & 95.29 \\  \midrule  
      NIPCC & 30.65 & 48.99 & 17.37 & 31.18 \\
      \bottomrule
      \end{tabular}}
  \end{center}
\end{table}

Analyzing the performance metrics of these models provides further insights (Table \ref{tab:CF_NIPCC}). The initial round of \judgeClimate{} (A) shows moderate precision (54.59\%) and high recall (81.01\%), leading to an F1 score of 50.90\% and accuracy of 83.53\%. After the second round, there is a notable improvement, with precision, recall, F1 score, and accuracy all increasing significantly. In contrast, \judgeClimate{} (B) starts with lower precision (36.99\%) and recall (70.74\%), resulting in an F1 score of 26.40\% and an accuracy of 58.24\%. The second round, however, brings a substantial enhancement, aligning its performance closely with \judgeClimate{} (A) in the second round and the original \judgeClimate{} model.

The consistent performance convergence in the second round across both variants of \judgeClimate{} highlights the robustness of the models in reassessing the claims. It suggests that despite the initial influence of the climate science denialist NIPCC \advocate{}, the iterative process enables a more comprehensive and evidence-based evaluation, ultimately aligning more closely with the scientific consensus as represented in the original \judgeClimate{} model.

\paragraph{Analyzing NIPCC's executive summary.}

In an endeavor to understand the narrative of the climate science denialist discourse, we turned our focus to the executive summary of the NIPCC's report \citep{idso2013climate}. The executive summary is particularly insightful as it encapsulates the key positions of the NIPCC on various climate-related issues. By extracting claims from this summary, we aimed to directly engage with the core statements presented by the NIPCC. This approach also allows for a nuanced exploration of how these claims are interpreted and validated by different models, especially in the context of climate science deniers. A typical example of such a claim is given below.
\begin{tcolorbox}[colback=blue!15!white,colframe=blue!15!white]\scriptsize \sffamily 
There has been no increase in the frequency or intensity of drought in the modern era. Rising CO2 lets plants use water more efficiently, helping them overcome stressful conditions imposed by drought.\\
  \rightline{\textit{\underline{source}}: NIPCC, Climate Change Reconsidered}
\end{tcolorbox}

Such claims, which mix false (frequency and intensity of droughts) and indisputable (CO2 and water efficiency of plants) elements, are typical of climate denier rhetoric. This tactic can cause confusion and obscure the overall understanding of climate change impacts. It underscores the importance of thorough and nuanced analysis when evaluating and categorizing climate-related claims. 

For the above claim, \judgeClimate{}, even when we include the NIPCC in the pool of advocates, would ask follow-up questions and would come to the following verdict after the second round of debate:

\begin{tcolorbox}[colback=orange!15!white,colframe=orange!15!white]\scriptsize
[...] The final assessment is that the claim about drought frequency and intensity is [[incorrect]], as authoritative sources indicate an increase in drought severity and intensity due to climate change. However, the claim that rising CO2 improves plant water use efficiency is [[mostly accurate]], acknowledging the CO2 fertilization effect. It is important to note that while CO2 can enhance plant water use efficiency, this does not fully mitigate the increased drought conditions driven by climate change.
\end{tcolorbox}


\begin{table}[h]
\centering
  \captionsetup{font=small}
\caption{\small Verdicts for Climate Change Reconsidered. The table reports the distribution of the granular verdicts of the NIPCC Advocate and \judgeClimate{} (for the first and second round of debate).} 
\label{tab:matrixCR}\scalebox{0.9}{
\begin{tabular}{lccc}
\toprule 
Verdict & NIPCC & \multicolumn{2}{c}{\judgeClimate}  \\
  &   & Round 1 & Round 2  \\
\midrule
incorrect & 3 & 46 & 46 \\ 
imprecise & 1 & 0 & 0 \\ 
misleading & 6 & 6 & 6 \\ 
mostly inaccurate & 0 & 0 & 1 \\ 
unsupported & 3 & 9 & 10 \\ 
lacks context & 6 & 5 & 5 \\ 
mostly accurate & 4 & 11 & 11 \\ 
correct & 0 & 2 & 2 \\ 
follow-up question & 0 & 2 & 0 \\  \midrule
\textsc{NEI} & 58 & 0 & 0 \\
\bottomrule
\end{tabular}}
\end{table}

Given the subtleties of these deniers' claims as exemplified above, we use a high granularity to analyze the claims. Table \ref{tab:matrixCR}) reports our results. A striking observation emerges. The NIPCC model, even when solely assessing claims from its own executive summary, exhibits a puzzling struggle in verifying these claims. Out of 81 NIPCC claims, it initially classified none of the claims from its own executive summary as correct, and 58 as \textsc{NEI}s, indicating a tendency to affirm more claims from its own report. This finding is intriguing and somewhat paradoxical, but also revealing about the missing soundness of climate skepticism. One might expect that a model based on the NIPCC’s own literature would readily affirm the claims it presents. However, the results suggest a more complex reality where the model's verification process does not uniformly align with the NIPCC's assertions. 

As expected, \judgeClimate{} emerges as a more discerning and accurate tool in the analysis of climate change claims. As shown in Table \ref{tab:matrixCR}, in the first round of assessment, \judgeClimate{} classified a significant number of claims as 'incorrect' (46 out of 81), starkly contrasting the NIPCC model's inclination towards \textsc{NEI} classifications. This outcome underlines \judgeClimate{}'s proficiency in critically evaluating claims against established scientific understanding and evidence. Moreover, the iterative process employed by \judgeClimate{}, involving rounds of debate and follow-up questions, enhances the model's capability to provide a more rounded and thorough analysis.

\section*{Conclusion}

Marvin Minsky's "The Society of Mind" \cite{minsky1986society} provides a compelling analogy for understanding the collaborative and iterative nature of modern AI systems in tackling complex challenges, such as those faced by LLMs. Through their collective intelligence, LLMs effectively address the issue of 'hallucination' in AI responses \cite{du2023improving, zhuge2023mindstorms}. The ‘agents’ conceptualized by Minsky find a parallel in the Advocate components of our \judgeClimate{} framework, each specializing in distinct aspects of climate science. This allows for a nuanced and multifaceted analysis of climate-related claims, significantly reducing biases and errors. 


Our application of \judgeClimate{} to the controversial statement by Sultan Al Jaber (see Appendix \ref{app:jaber}) exemplifies the framework's potential for providing transparent and objective evaluations. 
However, while \judgeClimate{} marks a significant step forward, it is not without limitations. Continuous updates to its database are essential to keep abreast of the latest scientific developments. Additionally, ongoing efforts to mitigate inherent biases and improve the model's accuracy are crucial for maintaining trust in AI-driven fact-checking tools. In conclusion, the \judgeClimate{} framework, not only offers a promising approach to fact-checking but also serves as a compelling argument about the effectiveness of AI in enhancing our understanding of complex global challenges.


\section*{Related Literature}

\judgeClimate{} is connected to the automated fact-checking and claim verification literature \cite{vlachos-riedel-2014-fact, guo-etal-2022-survey}, the fact-checking literature about environmental claims \cite{leippold2020climatefever}, and the interpretable fact checking literature \cite{fajcikClaimDissectorInterpretableFactChecking2023,atanasova-etal-2020-generating-fact, stammbach-ash, kotonya-toni-2020-explainable}. Automated fact-checking aims to find evidence for a given claim from a trustworthy knowledge base and determines whether the knowledge base supports or refutes the claim. Interpretable fact-checking approaches generate an interpretable explanation, given the claim, retrieved evidence, and predicted verdict. However, only a small amount of literature awares the complexity in claim verification regarding comprehensive reasoning \citep{chenComplexClaimVerification2023,jiangHoVerDatasetManyHop2020} and the controversial of evidence \citep{waddenSciFactOpenOpendomainScientific2022}. To our best knowledge, there is no existing work addressing these controversies of real-world claim verification. \judgeClimate{} studies this under-explored direction via aggregating explanations for different viewpoints with LLMs. 

Arguably, the individual viewpoints are essential to arrive at the ultimate verdict -- this approach is rooted in the deliberation literature \cite{Mercier2011, karadzhov2023delidata}. More modern interpretations of the automated fact-checking trend towards such a deliberation setting \cite{fact-checking-conversation} and incorporate generating questions and answers given a claim and trustworthy evidence to arrive at a fact check \cite{schlichtkrull2023averitec}. Such methods are also implemented in real life, e.g., in the \citet{community-notes} community notes. These notes provide additional context for a given Tweet, and if enough contributors from different points of view assess it as being useful, the Tweet receives such a note. One way to think about \judgeClimate{} is to consider the \advocates{} as contributors writing such notes.

Lastly, \judgeClimate{} relies on capable LLMs as its backbone \cite{openai2023GPT4} and retrieval-augmented language modeling \cite{gaoRetrievalAugmentedGenerationLarge2024,knn-lms, retro}, which can be used to integrate trustworthy and up-to-date information about climate change into these models \citep[see e.g.,][]{vaghefi2023chatclimate}. It is also in line with the work investigating the effectiveness of AI system aggregation \citep{chenSelfPlayFineTuningConverts2024,bansalLLMAugmentedLLMs2024}.


\section*{Limitations and Future Research}

While \judgeClimate{} represents a significant advancement in AI-supported climate discourse, it is subject to certain limitations that open avenues for future research and development.

\paragraph{Recency of information.} One of the key challenges lies in the tool's reliance on potentially outdated information. The rapidly evolving nature of climate science means that newer studies and findings might not be immediately integrated into the system's knowledge base.  To address this issue, future iterations of \judgeClimate{} could include a mechanism to prioritize and incorporate the most recent scientific publications and news, ensuring that the tool's assessments are always grounded in the latest information.

\paragraph{Source quality and comprehensiveness.} Another limitation is the scope and quality of sources from which \judgeClimate{} retrieves information. While the tool currently utilizes reputable sources, it may not capture the entire spectrum of relevant and emerging scientific data, especially if such information is behind paywalls. Hence, enhancing the information retrieval process and broadening the base of sources, including more diverse scientific journals, reports, and datasets, will be crucial. This will allow the tool to provide more comprehensive and robust answers to a wider array of climate-related queries.

\paragraph{Technical infrastructure.} Currently, \judgeClimate{} is built on GPT-4, a closed-source model, which might pose challenges in terms of transparency and consistency over time as the model evolves. Therefore, transitioning to open-source language model checkpoints that are specifically trained for climate science fact-checking tasks can offer more transparency and control. This shift would also facilitate continuous updates and customizations tailored to the evolving needs of climate science communication.

\paragraph{Multi-modal debating system.} Looking ahead, there is potential for expanding \judgeClimate{} into a multi-modal debating system. This would involve integrating various forms of data input, such as images, graphs, and videos, along with textual information. Such a system could provide a more holistic and interactive platform for climate science communication, engaging users in a richer and more informative experience. This multi-modal approach could significantly enhance the tool's ability to debunk myths and clarify misconceptions by presenting evidence in various formats, making the information more accessible and comprehensible to a broader audience.

\paragraph{Output Evaluation.} Lastly, extending our \judgeClimate{} framework to incorporate the evaluation approach presented in \citet{bulian2023assessing} could significantly enhance our understanding of the system's capabilities in climate communication. By adopting their comprehensive evaluation criteria focused on presentational and epistemological adequacy, we can conduct a more nuanced analysis of \judgeClimate{}'s outputs, ensuring both accuracy and clarity in climate-related information dissemination. This extension would allow us to benchmark our system against other LLMs, utilizing a scalable oversight protocol with AI assistance and educated raters. Such a rigorous analysis could highlight potential areas for improvement, contributing to the development of more reliable and effective AI tools in the realm of climate change communication and fact-checking.

\bibliography{literature}

\section*{Acknowledgements} This paper has received funding from the Swiss
National Science Foundation (SNSF) under the project `How sustainable is sustainable finance? Impact evaluation and automated greenwashing detection' (Grant Agreement No. 100018\_207800).



\section*{Additional information}

\textbf{Competing interests} None of the authors has competing interests. 

\appendix

\section*{Appendix}
\setcounter{footnote}{0}
In this Appendix, we provide supplementary material for our study.  

\section{Data Availability}
\label{sec:data}
The main data sources used in this study for RAG are:
\paragraph{IPCC:}
\begin{enumerate}
    \item Summary for Policymakers from each of the Working Groups (I, II, III): i 3 pdfs
    \item All chapters (WG I: Chapters 1-12, WG II: Chapters 1-18, Cross-Chapters 1-12, WGIII: Chapters 1-17) and Technical Summary from each of the three working groups. \footnote{\url{https://www.ipcc.ch/report/ar6/wg1/}} \footnote{\url{https://www.ipcc.ch/report/ar6/wg2/}} \footnote{\url{https://www.ipcc.ch/report/ar6/wg3/}}
    \item The IPCC Synthesis Report 2023: 1 pdf file. 
\end{enumerate}

\paragraph{WMO:}
\begin{enumerate}
    \item 2022 State of Climate Services: Energy (WMO No. 130) \footnote{\url{https://library.wmo.int/records/item/58116-2022-state-of-climate-services-energy}}
    \item WMO Global Annual to Decadal Climate Update, Target years: 2023 and 2023-2027 \footnote{\url{https://library.wmo.int/records/item/66224-wmo-global-annual-to-decadal-climate-update}}
    \item State of the Climate in Asia 2022 (WMO No. 1321) \footnote{\url{https://library.wmo.int/records/item/66314-state-of-the-climate-in-asia-2022?language_id=&offset=9}}
    \item State of the Climate in South-West Pacific 2021 (WMO No. 1302) \footnote{\url{https://library.wmo.int/records/item/66342-state-of-the-climate-in-south-west-pacific-2022}}
    \item State of the Climate in Africa 2021 (WMO No. 1300) \footnote{\url{https://library.wmo.int/records/item/58070-state-of-the-climate-in-africa-2021?offset=234}}
\end{enumerate}

\paragraph{Abstracts of Scientific Literature on Climate Change (AbsCC):}
abstracts of scientific literature indexed on the Web of Science from 2015 to 2022 that have 'climate change' in their titles, abstracts, and keywords in total 290k abstracts.

\paragraph{Abstracts of 1000Scientists (S1000):}
The Reuters list of the world’s top climate scientists was released in 2020. To read more about how Reuters built it, refer to their website\footnote{\url{https://www.reuters.com/investigates/special-report/climate-change-scientists-list/}}. Reuters created a system of identifying and ranking 1,000 climate academics according to how influential they are in the community. In this paper, we only used the climate-related Abstracts of scientists from 2015 to 2022, a total of 190k abstracts.

\paragraph{NIPCC:} The NIPCC, or Non-Governmental International Panel on Climate Change, is an international panel of scientists who are more skeptical stance. Their publications\footnote{\url{https://climatechangereconsidered.org/wp-content/uploads/2019/01/CCR-II-Physical-Science-10-17-2013-entire-book.pdf}}\footnote{\url{https://climatechangereconsidered.org/wp-content/uploads/2019/01/Full-Book.pdf}} often emphasize natural factors in climate change and criticize the methodologies and conclusions of the IPCC.

We restrict our study to these text sources. These sources could be further extended, and they could also be updated on a regular basis so as to have the most recent data available for fact-checking.

\section{Complexity analysis of climate claims}

\renewcommand{\thefigure}{\thesection\arabic{figure}}
\setcounter{table}{0}
\renewcommand{\thetable}{\thesection\arabic{table}}

\label{app:complex}
We analyze the semantic and syntactic differences of the claims from the different sources available. In particular, we quantify the average syntactic complexity of a claim by computing the mean number of tokens within the syntactic trees of all its sentences, serving as an indicator of sentence intricacy. We measure semantic complexity by first obtaining BERT embeddings for each claim, standardizing these embeddings, applying t-SNE for dimensionality reduction to two components,\footnote{We favor t-SNE over PCA for preserving local structures within high-dimensional data, effectively revealing intricate clusters and relationships that PCA, a linear method, might overlook.} and then calculating the Euclidean norm of these components, which serves as a proxy for the semantic intricacy of the claims.

The claims from Skeptical Science are characterized by a broad semantic spectrum, which indicates a wide array of topics pertinent to climate change. This extensive range may cover elements from scientific data and policy debates to the impacts on ecosystems. Conversely, the syntactic structure in Skeptical Science is noticeably less complex. The straightforward manner in which claims are articulated implies an effort to enhance accessibility. The emphasis on clarity and the minimization of linguistic complexity point towards a strategy aimed at demystifying climate change discourse, making it more approachable for non-specialist readers, which is indeed the goal of Skeptical Science.

In contrast, the set of claims from Climate Feedback demonstrates significant variation not only in semantic breadth but also in syntactic complexity. The content here oscillates between specialized, in-depth discussions and more generalized information, touching on nuanced perspectives on climate change. The semantic complexity, though substantial, does not reach the peaks observed for Skeptical Science, suggesting a more focused yet still comprehensive approach to topic selection. 

The elevated syntactic complexity observed in some segments of claims from Climate Feedback may indicate an academic or technical orientation. This complexity, often manifesting through formal and intricate sentence constructions, likely targets a reader base with a robust understanding of scientific and policy-related nuances. The sophisticated use of language aligns with the conventions of scholarly discourse, pointing towards an audience well-versed in the technicalities and subtleties of climate science.

\begin{figure}[ht]
\centering
\includegraphics[width=\linewidth]{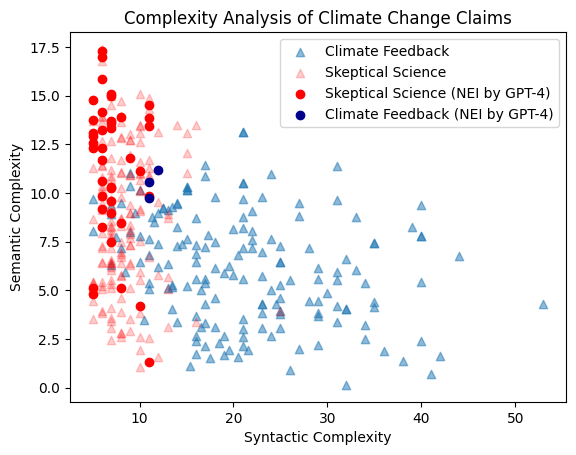}
  \captionsetup{font=small}
    \caption{The ratio of `Not Enough Information' (\textsc{NEI}) for different models. The model generates a \textsc{NEI} if it cannot access the information needed to provide a verdict on the claim taken from Skeptical Science.}
\label{fig:complexity}
\end{figure}

In summary, the higher incidence of \textsc{NEI}-classifications for Skeptical Science might be due to its broad but shallow approach, which contrasts with the type of detailed, context-rich texts that NLP models, including GPT-4, are generally more adept at processing. The lower syntactic complexity, while beneficial for human readability, might inadvertently strip away the contextual or semantic depth needed for a model to make a confident assessment.

\section{Analysis of a specific claim}\label{sec:amazon}
\setcounter{figure}{0}

The following claims led to a discrepancy between the assessment by Climate Feedback and \judgeClimate:

\begin{enumerate}
\item We have been cooling down for the past 4000 years; the Earth has cooled since the ‘medieval warming’, It’s all about when you start the measurements. (Climate Feedback: incorrect; \medClimate: correct)
\item Amazon near tipping point of switching from rainforest to savannah.
(Climate Feedback: incorrect; \medClimate: correct)
\item Surface air temperatures at the bottom of the world have risen three times faster than the global average since the 1990s.
(Climate Feedback: correct; \medClimate: incorrect)
\item The consensus among scientists and policy-makers is that we’ll pass this point of no return if the global mean temperature rises by more than two degrees Celsius
(Climate Feedback: incorrect; \medClimate: correct)
\item Sea level rise could reach six or seven feet by the year 2100.
(Climate Feedback: correct; \medClimate: incorrect)
\end{enumerate}

In the main text, we mentioned the claim ``Amazon near tipping point of switching from rainforest to savannah''. Here,  we further present the output of each \advocate{} and the \moderator, leading to the final verdict. Note that all of the following text, including the references, is generated by \judgeClimate, not only for this claim but for all the 413 claims analyzed in this paper.

\subsection{Response of IPCC} 
The claim ``Amazon near tipping point of switching from rainforest to savannah'' is supported by multiple sources cited in the provided information, which discuss the potential for a biome shift in the Amazon due to various factors including deforestation, climate change, and fire. The evidence and agreement levels vary across different aspects of this claim, but overall, there is a consistent concern about the Amazon approaching a tipping point.
\begin{enumerate}
\item The IPCC reports indicate that deforestation and climate change could lead to a tipping point resulting in the savannisation of the Amazon rainforest (IPCC AR6 WGII CCP7, p.12; IPCC AR6 WGII Chapter12, p.18). This is supported by evidence of already lost forest cover and potential future reductions in precipitation (IPCC AR6 WGII CCP7, p.12). 
\item The synergistic effects of deforestation, fire, and climate change exacerbate the risk of savannisation (IPCC AR6 WGII Chapter12, p.18), with medium confidence and medium evidence but high agreement.
\item Vegetation model simulations suggest a transition towards savanna-like vegetation, especially in the eastern Amazonian region, depending on the CO2 fertilisation effect and future dry season length (IPCC AR6 WGII CCP7, p.11).
\item The risk of a biome shift to grassland is increased by the lower buffering capacity for plant moisture during drought, combined with increased heat and fire (IPCC AR6 WGII Chapter02, p.76).
\item CMIP6 projections suggest that a tipping point may be crossed on a local or regional scale, but the timing and probability of an abrupt shift are difficult to ascertain (IPCC AR6 WGI Chapter08, p.96).
\item Droughts have already increased tree mortality in the Amazon, and the forest is becoming less of a carbon sink (IPCC AR6 WGII Chapter12, p.18).
\item The Amazon has become a net carbon emitter due to deforestation and associated burning (IPCC AR6 WGII Chapter02, p.55).
\item The Amazon ecosystem is listed as one of the IPCC Reasons for Concern due to climate change, with high confidence in the significant risks posed by warming and GHG emissions (IPCC AR6 WGII Chapter08, p.50).
\item Most simulations predict a significant geographical shift of transition areas between tropical forests and savanna, but forest dieback does not occur in the majority of simulations (IPCC AR6 WGII Chapter02, p.69).
\item The Amazon Forest has already shown increases in tree mortality and a decline in the carbon sink, potentially becoming a net carbon source (IPCC AR6 WGII Chapter16, p.89).
\item Wildfire-induced conversion of the Amazon rainforest to grassland is projected at temperature increases of 3°C-4.5°C (IPCC AR6 WGII Chapter02, p.85).
\item Deforestation, drying, and increased fire are projected to increase, potentially leading to a 50\% loss in forest cover by 2050 in a worst-case scenario (IPCC AR6 WGI Chapter08, p.95).
\item Climate models project drying in the Amazon, with increased burned area and higher risk of savannisation at higher GSAT levels (IPCC AR6 WGII Chapter16, p.90).
\item Continued anthropogenic disturbance and global warming may result in significant dieback of forest in the Amazon (IPCC AR6 WGII Chapter12, p.9).
\end{enumerate}
In summary, there is a high level of evidence and agreement that the Amazon is at risk of a significant biome shift from rainforest to savannah due to the combined effects of deforestation, climate change, and other factors. While the exact timing and extent of this shift are uncertain, the overall trend and potential for reaching a tipping point are well-supported by the scientific literature cited.\\
Level of evidence and agreement: High\\ 
\paragraph{Verdict:} The claim is {\bf mostly accurate}.

\subsection{Response of WMO} 
Based on the information provided:
\begin{enumerate}
    \item  The IPCC AR6 WG2 report is cited as stating that there is high confidence in increased drought severity and intensity, and a decline in soil moisture in specific regions, including the Amazon Basin, due to reduced precipitation and increased evaporative demand (State of the Climate in South-West Pacific 2021, Page 25, WMO). This supports the claim that the Amazon is expected to become drier.
    
    \item The WMO Global Annual to Decadal Climate Update for 2023-2027 predicts reduced rainfall over the Amazon for the May to September average relative to the 1991-2020 average (WMO Global Annual to Decadal Climate Update 2023-2027, Page 2, WMO). This prediction aligns with the claim of the Amazon becoming drier.
    
    \item However, the claim that the Amazon is near a tipping point of switching from a rainforest to savannah is not directly supported by the information provided. While the provided sources indicate a trend towards drier conditions, they do not explicitly state that the Amazon is near a tipping point or that a biome shift to savannah is imminent.
    
    \item The concept of a tipping point where the Amazon could transition from rainforest to savannah is a topic of scientific research and debate, but the specific information required to confirm the claim is not present in the provided references.
    \end{enumerate}
    In conclusion, the information supports the claim that the Amazon is expected to become drier, but there is not enough information to assess the claim that the Amazon is near a tipping point of switching from rainforest to savannah. Therefore, the claim is "Not Enough Information" to be fully evaluated.\\
    \paragraph{Verdict:} "Not Enough Information" 

\subsection{Response of AbsCC}
The claim that the Amazon is near a tipping point of switching from rainforest to savannah is supported by multiple sources provided, which discuss the risks and mechanisms that could lead to such a transition. Here is the evidence from the provided references:
\begin{enumerate}
    \item The commentary by Pereira et al., 2019, suggests that the resilience of the Amazon rainforest is jeopardized by climate and deforestation impacts, and that a tipping point leading to savannization is alarmingly conceivable (Reference: Catastrophic Climate Risk and Brazilian Amazonian Politics and Policies: A New Research Agenda, Page:1).
    \item Nobre et al., 2016, establish that there are two potential tipping points for the Amazon: a temperature increase of 4 °C or deforestation exceeding 40\% of the forest area, which could lead to large-scale savannization (Reference: Land-use and climate change risks in the Amazon and the need of a novel sustainable development paradigm, Page:1).
    \item Staal et al., 2015, indicate that the synergistic effects of precipitation reduction and deforestation could trigger regime shifts from forest to savanna in the south-eastern Amazon (Reference: Synergistic effects of drought and deforestation on the resilience of the south-eastern Amazon rainforest, Page:1). 
    \item De Faria et al., 2021, note that increased fire frequency and intensity could push large Amazon forest areas towards a tipping point, causing transitions to states with low tree and high grass cover (Reference: Climate change and deforestation increase the vulnerability of Amazonian forests to post‐fire grass invasion, Page:1). 
    \item Feeley et al., 2016, discuss the potential dangers of climate change for Amazonian parks, showing that a significant percentage of protected areas may not have future temperature analogs, which could affect the efficacy of these areas in conserving the rainforest (Reference: Disappearing climates will limit the efficacy of Amazonian protected areas, Page:1).
    \item The research by Rodriguez-Cordero et al., 2022, provides a contrasting perspective, predicting that Amazon moist forests will be relatively well-preserved under most future climate change scenarios (Reference: Predicting habitat suitability of Caiman yacare and assessing the role of protected areas under current and future climate and deforestation models, Page:1).
    \item The majority of the evidence points towards the Amazon being at risk of transitioning to a savanna due to the combined effects of climate change and deforestation. However, there is also mention of the potential for conservation and sustainable development to mitigate these risks. The claim is therefore supported by high evidence and high agreement among the sources, but it is important to note that the transition is not yet a certainty and depends on future actions regarding climate change and deforestation.
    \end{enumerate}
    Given the evidence, the overall assessment of the claim is that it is {\bf mostly accurate}, with the caveat that while the risk is high, the outcome is not yet determined and can be influenced by future policies and actions.

\subsection{Response of S1000} 
The provided references consistently support the claim that the capacity of Amazonian forests to sequester carbon has weakened, which has important implications for climate change. The references from Reuters articles and scientific papers provide evidence of long-term decline in the Amazon carbon sink (Reference: Long-term decline of the Amazon carbon sink, Page: 1), challenges in measuring changes in Amazon vegetation due to cloud cover and aerosols (Reference: On the measurability of change in Amazon vegetation from MODIS, Page: 1), and the threats to the Amazon forest from development decisions (Reference: Environmental policy in Brazilian Amazonia: Lessons from recent history, Page: 1). Additionally, the references discuss the historical stability and past expansions of savannah and grassland in the Amazon (Reference: Dry corridors opened by fire and low CO2 in Amazonian rainforest during the Last Glacial Maximum, Page: 1), the observed "drier dry season" and its expected intensification due to climate change (Reference: Simultaneous Regional Detection of Land‐Use Changes and Elevated GHG Levels: The Case of Spring Precipitation in Tropical South America, Page: 1), and the dangers of droughts, floods, and forest loss for the Amazon region (Reference: Drought, Floods, Climate Change, and Forest Loss in the Amazon Region: A Present and Future Danger?, Page: 1). The impact of climate variability and trends, land use change, and warming trends in the Amazon are also covered, highlighting the importance of the forest in climate and carbon cycles, the risk of fires, and the potential "die back" of the Amazon forests (Reference: Changes in Climate and Land Use Over the Amazon Region: Current and Future Variability and Trends, Page: 1). The role of the Guiana Shield rainforests in South American climate regulation and the potential consequences of deforestation are also discussed (Reference: The Guiana Shield rainforests—overlooked guardians of South American climate, Page: 1). \\ Furthermore, the increasing human pressure on the Amazon rainforest in Brazil, including the impact of infrastructure projects like Highway BR-319, is highlighted (Reference: Amazonian indigenous peoples are threatened by Brazil’s Highway BR-319, Page: 1). The significance of Southern Amazonia\'s carbon stocks for global climate and the impact of deforestation on greenhouse gas emissions are also reviewed (Reference: Brazil’s Amazonian forest carbon: the key to Southern Amazonia’s significance for global climate, Page: 1).\\ The encroachment of forest into savanna at the "Cerrado"–Amazon ecotone and the structural changes in tree species composition are documented (Reference: Savanna turning into forest: concerted vegetation change at the ecotone between the Amazon and “Cerrado” biomes, Page: 1). Lastly, the response of Amazon forests to severe droughts and the potential for recovery after water stresses are assessed using satellite remote sensing data (Reference: Amazon Forests’ Response to Droughts: A Perspective from the MAIAC Product, Page: 1).\\ Based on the information provided, the claim that the Amazon is nearing a tipping point of switching from rainforest to savannah is supported by the evidence of weakened carbon sequestration, increased human pressures, and environmental changes that could lead to significant biome shifts. The references collectively provide high evidence and high agreement for the claim.\\  {\bf Verdict: mostly accurate}.

\subsection{Response of GPT4} 
To assess the claim that the Amazon is near a tipping point of switching from rainforest to savannah, we need to consider the current scientific understanding of ecosystem tipping points, the specific research on the Amazon rainforest, and the impacts of deforestation and climate change on this ecosystem.
\begin{enumerate}
    \item  Ecosystem Tipping Points: A tipping point in an ecosystem is a threshold at which a relatively small change or disturbance in external conditions can lead to a rapid and often irreversible change in the state of the system. In the context of the Amazon, this could mean a shift from a moist, dense rainforest to a drier savannah-like ecosystem. This concept is well-established in ecological science. (In-house knowledge)
    \item Research on the Amazon Rainforest: There has been significant research suggesting that the Amazon rainforest could be approaching a tipping point due to the combined effects of deforestation, climate change, and fires. Studies have indicated that there is a threshold of deforestation beyond which the rainforest cannot sustain its own hydrological cycle. Once this threshold is crossed, the forest could transition to a savannah-like state. (In-house knowledge)
    \item Impacts of Deforestation and Climate Change: The Amazon has been experiencing high rates of deforestation, which reduces the forest's capacity to recycle water and maintain its microclimate. Climate change exacerbates this by increasing the risk of drought and fires, which can further degrade the forest. These factors are pushing the Amazon towards the tipping point. (In-house knowledge)
    \end{enumerate} 
    Given the scientific consensus on the existence of tipping points in ecosystems, the research indicating that the Amazon rainforest is under threat, and the observed impacts of deforestation and climate change, the claim that the Amazon is near a tipping point of switching from rainforest to savannah is supported by a substantial body of evidence. However, the exact proximity to this tipping point is difficult to quantify and may vary depending on the model and assumptions used. The level of evidence and agreement on the Amazon nearing a tipping point is medium to high. While there is strong evidence and a general scientific consensus that the Amazon is under threat and could reach a tipping point, predicting the exact timing or conditions for such a transition involves uncertainties. \\ 
    \paragraph{Verdict:} The claim is {\bf mostly accurate} based on current scientific understanding, but with the caveat that the complexity of the system and the uncertainties involved in predicting ecological tipping points mean that the exact proximity to this tipping point cannot be precisely determined.

\subsection{Assessment by the Mediator}
Review of Advocate Assessments:
\begin{enumerate}
    \item WMO Advocate: The WMO Advocate provides evidence of increased drought severity and intensity, and a decline in soil moisture in the Amazon Basin, supporting the claim that the Amazon is expected to become drier. However, it states that there is not enough information to assess the claim that the Amazon is near a tipping point of switching from rainforest to savannah. The verdict is "Not Enough Information."

\item IPCC Advocate: The IPCC Advocate supports the claim, citing evidence of deforestation, climate change, and fire increasing the risk of savannisation of the Amazon rainforest. It references various IPCC reports that discuss the potential for a biome shift and the factors contributing to this risk. The level of evidence and agreement is considered high, and the verdict is "mostly acurate."

\item AbsCC Advocate: This Advocate also supports the claim, referencing multiple sources that discuss the risks and mechanisms leading to a potential transition from rainforest to savannah. It acknowledges the high risk but also notes that the transition is not yet certain and can be influenced by future actions. The level of evidence and agreement is high, and the verdict is "mostly acurate."

\item S1000 Advocate: This Advocate supports the claim by providing evidence of long-term decline in the Amazon carbon sink, increased human pressures, and environmental changes that could lead to significant biome shifts. The references collectively provide high evidence and agreement for the claim. The verdict is "mostly accurate."

\end{enumerate}

\paragraph{Consolidation:}
The WMO Advocate's position is that there is not enough information to fully evaluate the claim about the tipping point, while the other three Advocates (IPCC, AbsCC, and S1000) provide evidence supporting the claim that the Amazon is at risk of a significant biome shift from rainforest to savannah. The IPCC, AbsCC, and S1000 Advocates all present a high level of evidence and agreement, which suggests a strong consensus on the risk of a biome shift.

\paragraph{Clarification:}
There is a discrepancy between the WMO Advocate's assessment and the other three Advocates. The WMO Advocate does not refute the claim but rather indicates a lack of specific information to confirm the tipping point. In contrast, the other Advocates provide specific evidence supporting the claim.

\paragraph{Final Assessment:}
Given the high level of evidence and agreement presented by the IPCC, AbsCC, and S1000 Advocates, and considering the WMO Advocate's position as not refuting but lacking specific information, the final verdict is that the claim is supported by the majority of the evidence provided. The WMO Advocate's caution is noted, but it does not outweigh the specific evidence provided by the other Advocates.

\paragraph{Final Verdict:} The claim is {\bf mostly accurate}.

\paragraph{User Query Summary:} The claim that the Amazon is near a tipping point of switching from rainforest to savannah is mostly accurate. While the WMO Advocate indicates a lack of specific information to confirm the tipping point, the IPCC, AbsCC, and S1000 Advocates provide high levels of evidence and agreement supporting the risk of a biome shift due to factors such as deforestation, climate change, and fire. The overall scientific consensus, as reflected in the provided documents, supports the claim that the Amazon is at risk of transitioning to a savannah, although the exact timing and extent of this shift are uncertain.

\section{Sultan Al Jaber's science claim}\label{app:jaber}

In a recent online event on November 21, Sultan Al Jaber, the president of Cop28 and chief executive of the United Arab Emirates’ state oil company Adnoc, made remarks that have sparked controversy among climate scientists and policymakers. Al Jaber claimed that there is no scientific evidence to suggest that a phase-out of fossil fuels is necessary to keep global heating within 1.5C, a statement that contradicts widely accepted scientific consensus. For this reason, we analyzed the claim ``There is no science out there, or no scenario out there, that says that the phase-out of fossil fuel is what's going to achieve 1.5C'' using \judgeClimate. Its verdict is ``incorrect'' and the reasoning based on the reasoning of the Mediator that aggregates the responses from the Advocates, given below.

\subsection*{Moderator's Reponse}

\paragraph{Review of Advocate Assessments:}

\begin{enumerate}
    \item WMO Advocate: The WMO Advocate provided evidence from multiple WMO reports that emphasize the importance of reducing greenhouse gas emissions, particularly from the energy sector, to meet the goals of the Paris Agreement. The reports indicate that achieving carbon neutrality and the 1.5°C target necessitates a significant reduction or phase-out of fossil fuel use. The level of evidence and agreement is high, and the verdict is that the claim is incorrect.

    \item IPCC Advocate: The IPCC Advocate cited various chapters of the IPCC reports that describe the necessity of phasing out fossil fuels to limit warming to 1.5°C. The reports detail the need for rapid penetration of renewable energy and a timely phasing out of fossil fuels, especially coal, to reach net zero CO2 emissions by around 2050/2060. The level of evidence and agreement is high, and the verdict is that the claim is incorrect.

    \item AbsCC Advocate: The AbsCC Advocate noted that while the phase-out of fossil fuels is a critical component of many scenarios to limit warming, it alone is not considered sufficient. Other measures such as carbon dioxide removal and a mix of policies are also deemed necessary. The level of evidence and agreement is high, and the verdict is that the claim is mostly accurate.

    \item S1000 Advocate: The S1000 Advocate provided references that consistently highlight the importance of reducing carbon dioxide emissions and transitioning to low-carbon technologies, including the phase-out of fossil fuels, as part of the strategy to limit global warming. The level of evidence and agreement is high, and the verdict is that the claim is incorrect.

\end{enumerate}

\paragraph{Consolidation and Clarification:} The WMO, IPCC, and S1000 Advocates all provide strong evidence that the phase-out of fossil fuels is a necessary component of achieving the 1.5°C target, directly contradicting the user's claim. The AbsCC Advocate also acknowledges the importance of fossil fuel phase-out but emphasizes that it must be accompanied by other measures, which does not refute the necessity of the phase-out but rather adds context to the complexity of the issue.

\paragraph{Final Assessment:} The evidence from the WMO, IPCC, and S1000 Advocates is highly consistent and authoritative, indicating that the phase-out of fossil fuels is supported by science and scenarios as a means to achieve the 1.5°C target. The AbsCC Advocate's assessment does not contradict this but adds that additional measures are also required, which aligns with the comprehensive approach to climate mitigation found in the literature.

Given the high level of evidence and agreement from the WMO, IPCC, and S1000 Advocates, and the additional context provided by the AbsCC Advocate, the final verdict is that the user's claim is \textbf{incorrect}. The phase-out of fossil fuels is indeed a scientifically supported strategy to achieve the 1.5°C target, although it is part of a broader set of actions needed for climate mitigation.

\paragraph{User Query Summary:}
The user claimed that there is no science or scenario that supports the phase-out of fossil fuels as a means to achieve the 1.5°C target. This claim has been assessed as incorrect based on the evidence provided by the WMO, IPCC, and S1000 Advocates, which all indicate that the phase-out of fossil fuels is a necessary component of the strategies to limit global warming to 1.5°C. The AbsCC Advocate adds that while the phase-out is critical, it must be part of a broader set of measures, which does not contradict the necessity of the phase-out.

\subsection*{Advocates' Reponses}
\subsubsection*{WMO}
The provided information from various WMO reports and IPCC assessments consistently emphasizes the need for a transition to low-carbon and renewable energy sources to meet the goals of the Paris Agreement and limit global warming to well below 2°C, with efforts to limit it further to 1.5 °C. The reports highlight the importance of achieving net zero carbon emissions by around 2050, the decarbonization of the energy sector, and the integration of low-and zero-carbon technologies. They also note the current inadequacy of pledges made by countries to meet these objectives and the significant gap in emissions reductions needed by 2030.

The claim that there is no science or scenario that supports the phase-out of fossil fuels to achieve the 1.5 °C target is directly contradicted by the information provided, which outlines the necessity of phasing out unabated coal, limiting investments in oil and gas, and embracing technology and policy solutions for a net zero pathway (Reference: 2022 State of Climate Services: Energy, WMO-No. 130, Page: 15, ORG: WMO, URL: \url{https://library.wmo.int/index.php?lvl=notice_display&id=22136#.Y3NxKOzMInd}.

Given the evidence from the provided references, the claim is [[incorrect]]. The level of evidence and agreement is high, as the information comes from authoritative and reputable sources such as the WMO and IPCC, which are based on extensive scientific research and assessments.']

\subsubsection*{IPCC}
The provided information from the IPCC reports consistently indicates that a rapid phase-out of fossil fuels and a significant increase in renewable energy are necessary to achieve the goals of the Paris Agreement, specifically limiting warming to 1.5°C with no or limited overshoot. The reports detail the required reductions in CO2 emissions, the need for net zero emissions by around 2050, and the significant decrease in the use of coal, oil, and natural gas to meet these targets. They also highlight the current insufficiency of national commitments (NDCs) to meet these goals and the urgency of immediate and substantial emissions reductions.

The claim that there is no science or scenario that supports the phase-out of fossil fuels to achieve the 1.5°C target is directly contradicted by the information provided in the IPCC reports. These reports represent the scientific consensus on climate change and are based on extensive research and analysis by the scientific community.

Given the evidence presented in the IPCC reports, the claim that the phase-out of fossil fuels is not necessary to achieve the 1.5°C target is [[incorrect]]. The level of evidence and agreement here is high, as the information comes from the authoritative and widely recognized Intergovernmental Panel on Climate Change.'

\subsubsection*{AbsCC}
The provided references collectively suggest that limiting global warming to 1.5 °C is a complex and challenging goal that requires immediate and ambitious action, including radical emissions cuts and the development of carbon dioxide removal technologies. The references also indicate that overshooting the 1.5 °C target could have irreversible consequences, and that many current scenarios may not be realistic or sufficient to achieve this goal. The economic implications are uncertain, but the cost of inaction could be high. The references also highlight the importance of integrating demand-side measures, technological innovation, and behavioral changes to meet the 1.5 °C target. They acknowledge the difficulties associated with large-scale carbon capture and storage (BECCS) and the competition for land with food production, suggesting the need for a diversified approach to mitigation pathways.

The claim that "there is no science out there, or no scenario out there, that says that the phase-out of fossil fuel is what\'s going to achieve 1.5C" is [[incorrect]]. The references provided clearly indicate that limiting global warming to 1.5 °C will require significant reductions in greenhouse gas emissions, which includes the phase-out of fossil fuels. For instance, the reference from Nogrady et al., 2021, explicitly states that most fossil-fuel reserves must remain untapped to hit the 1.5 °C warming goal, which implies that a phase-out of fossil fuels is a necessary component of achieving this target.

The level of evidence and agreement here is high, as the references consistently support the need for a phase-out of fossil fuels as part of the broader strategy to limit global warming to 1.5 °C.

\subsubsection*{1000Scientists}

The claim that there is no science or scenario that supports the phase-out of fossil fuels as a means to achieve the 1.5°C target is contradicted by the information provided. The references from Reuters consistently highlight the importance of a range of mitigation strategies, including the reduction of global energy demand, decarbonization of energy production, and the development of land-management systems, as well as the deployment of carbon dioxide removal technologies. These strategies inherently involve the phase-out of fossil fuels as a key component to limit global warming to 1.5°C, in line with the Paris Climate Agreement.

The repeated references to the necessity of supporting low-carbon technologies and phasing out the use of fossil fuels (Reference: Carbon dioxide emissions continue to grow amidst slowly emerging climate policies, Page: 1, URL: \url{https://doi.org/10.1038/s41558-019-0659-6}) directly imply that the scientific consensus acknowledges the role of fossil fuel reduction in achieving climate targets.

Furthermore, the detailed analysis of the IPCC 1.5°C special report scenarios (Reference: All options, not silver bullets, needed to limit global warming to 1.5 °C: a scenario appraisal, Page: 1, URL: https://doi.org/10.1088/1748-9326/abfeec) and the assessment of integrated assessment model scenarios (Reference: Energy system changes in 1.5°C, well below 2°C and 2°C scenarios, Page: 1, URL: \url{https://doi.org/10.1016/j.esr.2018.12.006}) both support the conclusion that significant changes in energy systems, which include moving away from fossil fuels, are necessary to meet the 1.5°C target.

Therefore, based on the information provided, the claim that there is no science or scenario that supports the phase-out of fossil fuels to achieve the 1.5°C target is [[incorrect]]. The level of evidence and agreement is high, as the claim directly contradicts multiple sources that are part of the scientific consensus on climate change mitigation strategies.

\clearpage
\onecolumn

\footnotesize
\renewcommand{\arraystretch}{1.5}

\section{Overview Climate Feedback claims}
\begin{longtable}{p{0.4\textwidth}p{0.15\textwidth}p{0.1\textwidth}p{0.15\textwidth}p{0.15\textwidth}}
\toprule
                                                                                           Claim & Source & Date & Climate Feedback &     \judgeClimate  \\
\midrule
\endfirsthead

\toprule
                                                                                           Claim & Source & Date & Climate Feedback &     \judgeClimate  \\
\midrule
\endhead
\midrule
\multicolumn{5}{r}{{Continued on next page}} \\
\midrule
\endfoot

\bottomrule
\endlastfoot
      Magnetic poles reversals involve the Earth flipping vertically and momentarily stopping its rotation, causing cataclysmic events during 6 days.  & Joe Rogan      \& Jimmy Corsetti, TikTok      \& Spotify & 14-Jun-23 & incorrect &incorrect \\
    “We have been cooling down for the past 4000 years”; the Earth has cooled since the ‘medieval warming’, “It’s all about when you start the measurements”  &  Ian Plimer, Media Research Center & 15-May-23 &inaccurate & mostly accurate \\
  Antarctic sea-ice extent is increasing to record levels. Antarctica is getting colder; its sea ice and ice cap are not melting  &   Judith Curry and Don Easterbrook, CFACT & 28-Apr-23 &inaccurate &      inaccurate \\
 Chemtrails contain toxic substances such as barium salts that are 10 thousand times more toxic to your nervous system than lead  &    William Deagle, Facebook & 06-Apr-23 &inaccurate &     unsupported \\
   Climate changed naturally in prehistoric eras, modern climate change is a naturally occurring phenomenon  &Social media user, Instagram & 23-Jan-23 & flawed reasoning &incorrect \\
 Scientists have found no consistent correlation between CO2 and temperature; scientists do not know whether earth will warm or cool in future  &Tom Harris, Fox News & 18-Oct-22 &inaccurate &incorrect \\
    “The question is, ‘how much of an impact do we have on climate?’ That has not totally been quantified”; climate change is happening anyway, “The ice age happened without us…It’s probably this constant cycle”.  &    Joe Rogan, Joe Rogan Experience & 11-Oct-22 & flawed reasoning &      inaccurate \\
CO2 is not the cause of our current warming trend; Arctic sea ice has expanded in recent years; Polar bears’ population is growing and is not threatened by climate change; CO2 is good for plant life  &     Patrick Moore, The Epoch Times & 20-Sep-22 & incorrect &incorrect \\
    “Believing CO2 controls the climate is pretty close to believing in magic.”  &   Richard Lindzen, Climate Change is Crap & 02-Sep-22 & incorrect &incorrect \\
   “Natural variation explains a substantial part of global warming observed since 1850;” no statistical evidence that global warming is intensifying natural disasters, or making them more frequent.  &   Toby Young, Daily Sceptic & 24-Aug-22 & incorrect &incorrect \\
      Magnetic poles reversals involve the Earth flipping vertically and momentarily stopping its rotation, causing cataclysmic events during 6 days.  & Joe Rogan      \& Jimmy Corsetti, TikTok      \& Spotify & 14-Jun-23 & incorrect &incorrect \\
    “We have been cooling down for the past 4000 years”; the Earth has cooled since the ‘medieval warming’, “It’s all about when you start the measurements”  &  Ian Plimer, Media Research Center & 15-May-23 &inaccurate &      inaccurate \\
  Antarctic sea-ice extent is increasing to record levels. Antarctica is getting colder; its sea ice and ice cap are not melting  &   Judith Curry and Don Easterbrook, CFACT & 28-Apr-23 &inaccurate &      inaccurate \\
 Chemtrails contain toxic substances such as barium salts that are 10 thousand times more toxic to your nervous system than lead  &    William Deagle, Facebook & 06-Apr-23 &inaccurate &     unsupported \\
   Climate changed naturally in prehistoric eras, modern climate change is a naturally occurring phenomenon  &Social media user, Instagram & 23-Jan-23 & flawed reasoning &incorrect \\
 Scientists have found no consistent correlation between CO2 and temperature; scientists do not know whether earth will warm or cool in future  &Tom Harris , Fox News & 18-Oct-22 &inaccurate &incorrect \\
    “The question is, ‘how much of an impact do we have on climate?’ That has not totally been quantified”; climate change is happening anyway, “The ice age happened without us…It’s probably this constant cycle”.  &    Joe Rogan, Joe Rogan Experience & 11-Oct-22 & flawed reasoning &      misleading \\
CO2 is not the cause of our current warming trend; Arctic sea ice has expanded in recent years; Polar bears’ population is growing and is not threatened by climate change; CO2 is good for plant life  &     Patrick Moore, The Epoch Times & 20-Sep-22 & incorrect &incorrect \\
    “Believing CO2 controls the climate is pretty close to believing in magic.”  &   Richard Lindzen, Climate Change is Crap & 02-Sep-22 & incorrect &incorrect \\
   “Natural variation explains a substantial part of global warming observed since 1850;” no statistical evidence that global warming is intensifying natural disasters, or making them more frequent.  &   Toby Young, Daily Sceptic & 24-Aug-22 & incorrect &incorrect \\
     Carbon dioxide is not a climate heat driver; we had an ice age when the Earth had extraordinarily larger amounts of carbon dioxide; in the 1970s scientists thought we were going to have a new ice age.  &      Jerome Corsi, CBN News & 05-Aug-22 & incorrect &incorrect \\
      Scientists have neglected the effect of solar particles, cosmic rays, the interplanetary magnetic field and Earth's weakening magnetic field to conclude climate change is due to human activity  &  Ben Davidson, Suspicious0bservers & 29-Jul-22 & incorrect &incorrect \\
    ”Global warming started to run out of steam over two decades ago… Half of the apparent global temperature increase from January 1910 to January 2000 is due to administrative adjustments.”  &      Chris Morrison, Daily Sceptic & 09-Jun-22 & incorrect &incorrect \\
     “...Seven and a half years have passed since there was any trend in global warming at all” If global warming was caused by human emissions, temperature would rise at a steady rate  &  Christopher Monckton, Watts Up With That & 11-Apr-22 & incorrect &incorrect \\
  There is no evidence, other than models, that human CO2 emissions drive climate change and abundant evidence that the Sun, coupled with natural climate cycles, drives most, if not all, of recent climate changes  &     Andy May, CO2 Coalition & 16-Mar-22 & incorrect &incorrect \\
 More than 75 percent of the rainforest is losing resilience; More than half of the rainforest could be converted into savanna in a matter of decades  &      Sarah Kaplan, Washington Post & 15-Mar-22 &  mostly accurate & mostly accurate \\
     Study Shows The Great Barrier Reef Is Growing Quickly  &Ben Zeisloft, The Daily Wire & 30-Oct-21 &misleading &incorrect \\
An electric car “costs more than seven times as much” as a gasoline powered car  &    Facebook users, Facebook & 21-Oct-21 &inaccurate &incorrect \\
 the US corn crop, at its peak, produces 40     \% more oxygen than the Amazon rainforest.  &  Facebook users, Facebook (see claim in context) & 23-Sep-21 &inaccurate &     unsupported \\
Global warming saves 166,000 lives each year; those claiming that climate change is causing heat-related deaths are wrong because they ignore that the population is growing and becoming older  &    Bjorn Lomborg, New York Post, Facebook & 15-Sep-21 &      unsupported &incorrect \\
      Great Barrier Reef Sea Surface Temperature: No Change In 150 Years  &David Mason-Jones, Wattsupwiththat? & 08-Sep-21 &inaccurate &incorrect \\
   The Sun and not human emissions of carbon dioxide may be the main cause of warmer temperatures in recent decades. There is a systemic bias in UN IPCC's data selection.  &     Alex Newman, The Epoch Times, The Western Journal, Newsmax & 30-Aug-21 & incorrect &incorrect \\
      “The amount of coral on the Great Barrier Reef is at record high levels”; those claiming the reef is threatened by climate change are alarmists  &     Peter Ridd, The Australian, Climate Change Dispatch, Watts Up With That? & 21-Aug-21 &inaccurate &incorrect \\
  The official observed temperatures are coming consistently below the scary global warming numbers; Melting polar ice triggers ice age  &  Ben Davidson, Suspicious0bservers & 27-May-21 &inaccurate &incorrect \\
 Plugging this device into a car’s fuse box can reduce fuel consumption by at least 35     \% and up to 75     \% by converting the car into a hybrid.  &      Anonymous, Christianity Today & 05-May-21 & incorrect &     unsupported \\
    Biden’s climate plan limits meat consumption by 90     \%; “Biden's climate plan could limit you to eat just one burger a MONTH”  &     Emily Crane, Daily Mail & 29-Apr-21 &inaccurate &      misleading \\
A third of Antarctic ice shelf risks collapse as our planet warms  &    Amy Woodyatt, CNN & 16-Apr-21 &  accurate & mostly accurate \\
  There is no factual data to back up global warming; real scientists don’t know whether CO2, solar sunspots or natural activity cause global warming  &  Candace Owens, Charlie Kirk , Turning Point USA & 16-Apr-21 &inaccurate &incorrect \\
   Deserts are heat producers, reflecting around 60     \% to 70     \% of the solar energy that falls on them straight back into the atmosphere.; If we want to do something about global warming, we have to do something about deserts.  &    Ties Van der Hoeven, Steve Rose, The Guardian & 09-Apr-21 &inaccurate &   lacks context \\
   Wind turbines in the US are made of toxic materials that pollute the environment, are mostly built in Germany, and will not pay off their manufacturing costs within their lifetimes.  &     Internet User, Facebook & 08-Apr-21 &inaccurate &      inaccurate \\
      The intentional dimming of direct sunlight by aircraft dispersed particles, a form of global warming mitigation known as “Solar Radiation Management”, has and is causing catastrophic damage to the planet’s life support systems.  &      Dane Wigington, Geoengineering Watch & 25-Mar-21 & incorrect &     unsupported \\
The rotating wheel of an electric vehicle can recharge the vehicle’s battery by affixing a generator  &   Ozzy's Classics, Facebook & 15-Mar-21 & incorrect &     unsupported \\
There exists a car that runs solely on water  &      Anonymous, Gaia & 15-Mar-21 &inaccurate &incorrect \\
 Government funded science: 70’s: another ice age in 10 years; 80’s: acid rain will destroy all crops in 10 years; 90’s: the ozone layer will be destroyed in 10 years; 2000’s the ice caps will be gone on 10 years  &    Facebook users, Facebook & 03-Mar-21 &inaccurate &      misleading \\
HAARP may deliberately create destruction on a global scale, causing earthquakes, cyclones, flooding, snowstorms, around the world; HAARP may be attacking the citizens of the world telepathically, influencing thoughts with low frequency vibrations  &      Anonymous, Gaia & 02-Mar-21 & incorrect &incorrect \\
    Helicopters are using chemicals made from fossil fuels to defrost wind turbines; frosted wind turbines are responsible for blackouts in Texas; “the reason your power is/was out is wind energy's fault.”  &    Facebook users, Facebook & 17-Feb-21 &inaccurate &      misleading \\
   “The 58 year net gain of global temperature is zero. We are simply in a cyclical, normal ebb and flow of temperature.”  &    Facebook users, Facebook & 05-Feb-21 &inaccurate &incorrect \\
climate models that attempt to predict the future temperature of the planet...don’t work. They haven’t worked in the past. They don’t work now.; the number of factors that influence climate—the sun, the earth’s orbital properties, oceans, clouds, and, yes, industrial man—is huge and enormously variable; CO2 is a minor contributor to the warming of the earth  &     William Happer, PragerU & 27-Jan-21 & incorrect &incorrect \\
   “there has not been a long-term distinctive change in sea level rise rates in the last 120 years”  &Pierre L. Gosselin, Kenneth Richard, NoTricksZone, Watts Up With That? & 15-Jan-21 &inaccurate &incorrect \\
A young whizkid ... invented a device that cuts 50     \% off an average home’s power bill  &    Anonymous, HOVERH & 27-Dec-20 &inaccurate &     unsupported \\
   “It’s a good thing that we are putting some more carbon dioxide in the atmosphere. The scientific method has not been applied in such a way as to prove that carbon dioxide is causing the Earth to warm.”  &    Patrick Moore, Instagram & 18-Dec-20 & incorrect &incorrect \\
 “The clothing industry contributes up to 10     \% of the pollution driving the climate crisis.”  & Patagonia, Instagram & 06-Dec-20 &  accurate & mostly accurate \\
      “The Earth is not a globe. The Earth is stationary.”  &   Amber Plaster, Flat Earth & 27-Nov-20 &inaccurate &incorrect \\
      “this is ... a brush rake. It's used for clearing underbrush and debris from the forest floor that builds up and creates fuel for wild fires. Now you can stop looking stupid trying to make fun of Trump telling California to rake the forest.”  &    Facebook users, Facebook & 04-Nov-20 &misleading &     unsupported \\
   “In 1950, there were around 10,000 polar bears globally. Today, polar bear populations are near 39,000. Polar bear populations are increasing dramatically as the planet has warmed.”  &Andy Singer, Climate Realism, Heartland Institute & 29-Oct-20 &misleading &      inaccurate \\
    “Amazon near tipping point of switching from rainforest to savannah”  &  Fiona Harvey, The Guardian & 14-Oct-20 & imprecise & mostly accurate \\
    “Forest fires are caused by poor management. Not by climate change.”  &  Anonymous, Facebook & 16-Sep-20 &misleading &incorrect \\
     a distressing number of the West coast fires are not accidents, and willful arson is the cause.  &Glen Morgan, We The Governed & 14-Sep-20 &inaccurate &     unsupported \\
     “Trees provide a solution to almost all environmental crises we're facing. Trees reverse climate change by cleaning carbon from the air.”  &   Pamela Anderson, 8 Billion Trees & 25-Aug-20 &    lacks context &   lacks context \\
   Global warming is driving polar bears toward extinction  & Henry Fountain, The New York Times & 27-Jul-20 &   correct & accurate \\
    Greenhouse gases emitted into the ocean are causing more holes in the ozone layer … “the ozone layer has holes in it causing global temperature to rise”  &     Otis Holland, Instagram & 21-Jul-20 & incorrect &incorrect \\
      Surface air temperatures at the bottom of the world have risen three times faster than the global average since the 1990s.  & Henry Fountain, The New York Times & 06-Jul-20 &  accurate &     unsupported \\
   The scientific consensus on whether global warming is human caused is not 97 percent, it’s less than 1 percent. There is no way to measure the human impact on climate change.  & Mark Mathis, Clear Energy Alliance & 25-Jun-20 & incorrect &incorrect \\
The sun has gone into ‘lockdown’ which could cause freezing weather, earthquakes and famine, say scientists  &      Chris Pollard, The Sun & 20-May-20 & incorrect &incorrect \\
    The polar bear population has been growing. Polar bears are thriving even where sea ice is diminishing.  &   Anonymous, PragerU & 18-May-20 & incorrect &incorrect \\
      The drop in NO2 content has only been “slight”. So it cannot be the evil diesel engine cars that are “choking” our cities.  &    Holger Douglas, Tichys Einblick & 02-May-20 & flawed reasoning &incorrect \\
    Global warming slowing down? 'Ironic' study finds more CO2 has slightly cooled the planet  &  Sebastian Kettley, Express & 24-Apr-20 & imprecise &incorrect \\
  Human additions of CO2 are in the margin of error of current measurements and the gradual increase in CO2 is mainly from oceans degassing as the planet slowly emerges from the last ice age.  &   Cole Ryan, I Love Carbon Dioxide & 20-Apr-20 & incorrect &incorrect \\
    They tell us that we are the primary forces controlling earth temperatures by the burning of fossil fuels and releasing their carbon dioxide. I hope my readers can recognize the absurdity of their claims. … increasing sunspots are linked to increases in earth temperature  &     Jay Lehr, CFACT  & 15-Apr-20 & incorrect &incorrect \\
    The Great Barrier Reef is experiencing the most widespread bleaching ever recorded  &  Michael Slezak and Penny Timms, ABC News & 11-Apr-20 &  accurate & accurate \\
    The ozone layer is healing  &      Louise Boyle, The Independent & 07-Apr-20 &  mostly accurate & mostly accurate \\
Deforestation has made humans more vulnerable to pandemics  &Lani Chan and Tom McKenna, Now This & 01-Apr-20 &  mostly accurate & mostly accurate \\
  The long-term tide gauge datasets are all in agreement that there is no acceleration  &    Willis Eschenbach, Watts Up With That? & 01-Apr-20 &inaccurate &incorrect \\
 Historical data of temperature in the U.S. destroys global warming myth  &     Cap Allon, Electroverse & 24-Mar-20 & flawed reasoning &incorrect \\
CO2 is a plant nutrient...it’s not a pollutant that threatens human civilization. If CO2 was so terrible for the planet, then installing a CO2 generator in a greenhouse would kill the plants.  &    Mike Adams, Natural News & 19-Mar-20 & flawed reasoning &      misleading \\
the IPCC is wrong - the sun, not CO2, drove modern global warming  &   Roger Higgs, Electroverse & 18-Mar-20 & incorrect &incorrect \\
      Sea level rise has been slow and a constant, pre-dating industrialization  &    Thomas Lifson, American Thinker & 12-Mar-20 &inaccurate &incorrect \\
    90 percent of the world’s glaciers are growing.  &    IceAgeNow, Principia Scientific International & 26-Feb-20 &inaccurate &incorrect \\
Of course the climate is changing. It always has. It always will.  &      Patrick Moore, PragerU & 21-Feb-20 &misleading &      misleading \\
Scientists were caught 'adjusting' sea level data to create false impression of rising oceans  &   Vicki Batts, James Delingpole, Signs of the Times, Breitbart & 20-Feb-20 &inaccurate &   lacks context \\
Using complex statistical models, NOAA changes the data to reflect not reality, but their underlying theories of global warming.  &      Anonymous, Investor's Business Daily & 19-Feb-20 &inaccurate &     unsupported \\
 Less than 5     \% of carbon dioxide emissions are produced by man ... they can’t cause 30     \% of today’s atmospheric carbon dioxide  &I     \&I Editorial Board, Issues      \& Insights & 11-Feb-20 &misleading &incorrect \\
 Earth about to enter 30-YEAR ‘Mini Ice Age’  &      Harry Pettit, Sean Martin, The Sun, Express & 06-Feb-20 & incorrect &incorrect \\
    Volcanoes Melting West Antarctic Glaciers, Not Global Warming  &    Robert W. Felix, James Edward Kamis, IceAgeNow.info, PlateClimatology.com & 20-Jan-20 &      unsupported &incorrect \\
  CO2 is not a greenhouse gas.  &  Tim Ball, Technocracy.news & 14-Jan-20 &inaccurate &incorrect \\
      The idea that 'climate change' is new or is caused by Humans is nonsense.  &  Anonymous, Facebook & 13-Jan-20 & incorrect &incorrect \\
the bushfires in Australia were caused by arsonists and a series of lightning strikes, not 'climate change'  &   Sean Adl-Tabatabai, Thomas D. Williams, NewsPunch, Breitbart & 08-Jan-20 &misleading &      misleading \\
   Discovery Of Massive Volcanic CO2 Emissions Discredits Global Warming Theory  &  James Edward Kamis, Principia Scientific & 03-Jan-20 &inaccurate &incorrect \\
      Father of Global Warming’ Scientist Finally Admits Theory Is Wrong  & Baxter Dmitry, NewsPunch, Your News Wire, Principia Scientific & 01-Jan-20 &inaccurate &incorrect \\
     Harvard study finds that wind turbines create MORE global warming than the fossil fuels they eliminate  &    Ethan Huff, Ecology.news & 23-Dec-19 &inaccurate &incorrect \\
 Photo A little science lesson for the      \#idiots at the global warming conference. Ice berg melts, ocean level remains the same.  &  Anonymous, Facebook & 17-Dec-19 & incorrect &   lacks context \\
 When the exact same group of 'experts' who claimed it was global cooling in 1977 now claim it's global warming you can easily see why I am skeptical  &  Anonymous, Facebook & 16-Dec-19 &inaccurate &incorrect \\
   A series of just-released studies by working-level scientists prove that geological and not atmospheric forces are responsible for melting of Earth’s polar ice sheets.  &      James Edward Kamis, Climate Change Dispatch & 22-Nov-19 &inaccurate &incorrect \\
    Scientists studying Antarctica sea ice warn a rise in accumulation could spark the next ice age.  &   Stacy Liberatore, Chris Ciaccia, Daily Mail, Fox News & 01-Nov-19 & incorrect &      misleading \\
 That one little burp by Mt. Etna has already put more than 10,000 times the CO2 into the atmosphere than mankind has in our entire time on the Earth  &    Anonymous, Facebook post & 31-Oct-19 &inaccurate &incorrect \\
 500 Scientists Write U.N.: ‘There Is No Climate Emergency’.... More than 500 scientists and professionals in climate and related fields have sent a 'European Climate Declaration' to the Secretary-General of the United Nations  &      Thomas D. Williams, Breitbart & 30-Oct-19 &misleading &     unsupported \\
Unprecedented climate change has caused sea level at Sydney Harbour to rise approximately 0.0 cm over the past 140 years.  &    Anonymous, Facebook post & 28-Oct-19 &inaccurate &incorrect \\
  The consensus among scientists and policy-makers is that we’ll pass this point of no return if the global mean temperature rises by more than two degrees Celsius  &   Jonathan Franzen, The New Yorker & 17-Sep-19 & incorrect & mostly accurate \\
      Human-produced carbon might be one of the factors of climate change, but there’s simply no evidence that it is a significant one.  &   Kurt Schlichter, Townhall & 11-Sep-19 & incorrect &incorrect \\
      Peer-reviewed studies, geologic records, and all the studies have shown that we have actually cooled since the Roman Warming Period, and likely since the Medieval Warming Period.  &    Marc Morano, TFP Student Action & 09-Sep-19 &inaccurate &incorrect \\
the U.S. Climate Reference Network... finds there has been no warming for the past 14 years at least  &  James Taylor, John Hinderaker, WND, Powerlineblog, Real Clear Energy & 06-Sep-19 &inaccurate &incorrect \\
      changes in the solar orbit of the earth, along with alterations to the earth’s axial tilt, are both responsible for what climate scientists today have dubbed as “warming”.... In no way, shape, or form are humans warming or cooling the planet  &    Ethan Huff, Natural News & 03-Sep-19 & incorrect &incorrect \\
      I am talking about the slaughter, death, and starvation of 6 billion people this century—that's what the science predicts.  &      Roger Hallam, BBC News & 22-Aug-19 & incorrect &     unsupported \\
     I think about all the 194 countries that signed onto the Paris accord, the U.S. is the one that's leading the world in reducing emissions  & Rick Perry, Fox News & 20-Aug-19 &misleading &      misleading \\
 CO2 does not cause climate change, it RESPONDS to it ... temperature always changes first, and CO2 follows  &     Robert W. Felix, William Happer, Ice Age Now & 14-Aug-19 & flawed reasoning &incorrect \\
  Based on the increase of solar activity during the twentieth century, it should account for between half to two-thirds of all climate change  &Cap Allon, Nir Shaviv, Electroverse & 13-Aug-19 & incorrect &incorrect \\
      The current solar minimum could last for more than three decades which could lead to temperatures plummeting across the globe, scientists have warned.  & Sean Martin, Express & 06-Aug-19 & incorrect &     unsupported \\
     18 proxies tell us the world was the same or warmer 1,000 years ago  &     Joanne Nova, joannenova.com.au & 05-Aug-19 &inaccurate &     unsupported \\
    The raw data, the actual thermometer data... shows that the US has been cooling for 80 to 90 years. But the graphs they release to the public show warming trend, and it's all because they've altered the data.  &    Tony Heller, EXOMATRIXTV & 05-Aug-19 &inaccurate &incorrect \\
   Earth’s mean temperature over the last 2,000 years shows two previous periods when temperatures were warmer than they are now; from 1–200 A.D., an epoch called the Roman Warm Period, and more recently the Medieval Warm Period from 900–1100 A.D.  &     Gregory Rummo, Townhall & 26-Jul-19 &inaccurate &incorrect \\
      the Intergovernmental Panel on Climate Change is misleading humanity about climate change and sea levels, and that in fact a new solar-driven cooling period is not far off  & Alan Jones, Nils Axel-Mörner, Sky News Australia & 18-Jul-19 & incorrect &incorrect \\
    During the last hundred years the temperature is increased about 0.1°C because of carbon dioxide. The human contribution was about 0.01°C.  &     Jyrki Kauppinen, Pekka Malmi, Paul Joseph Watson, Summit.news, Zerohedge & 12-Jul-19 & incorrect &incorrect \\
     So that means that probably about half, maybe half of that nine-tenths of the degree of total warming might be caused by greenhouse gases  &  Patrick Michaels, Fox News & 09-Jul-19 & incorrect &incorrect \\
  NASA has been adjusting temperatures from the past... downward, while adjusting current-day temperatures upward, and those changes are responsible for most of the claimed global warming during that time.  &      Tony Heller, James Murphy, The New American & 01-Jul-19 &inaccurate &     unsupported \\
  New Study Confirms EVs Considerably Worse For Climate Than Diesel Cars  &    Anonymous, Zero Hedge, Infowars & 24-Jun-19 &inaccurate &incorrect \\
     The UN Admits That The Paris Climate Deal Was A Fraud  &      Anonymous, Investor's Business Daily & 17-Jun-19 &      unsupported &incorrect \\
      A windmill could spin until it falls apart and never generate as much energy as was invested in building it.  &    Internet users, Facebook & 27-May-19 &inaccurate &incorrect \\
   the concentration of carbon dioxide in Earth’s atmosphere has climbed to a level last seen more than 3 million years ago — before humans even appeared on the rocky ball we call home  &Denise Chow, NBC News & 18-May-19 &  accurate & accurate \\
 Fifty-five thousand years ago the whole world was 2°C warmer than it is today... It's all cyclical; human-caused greenhouse gas emissions are not a fart in the wind. Sea level rise is not going to happen.  &      Dan Peña, "Ivanka     \_Trump" Facebook account & 07-May-19 & flawed reasoning &incorrect \\
      Climate change is not happening, there is no significant man-made global warming now, there hasn't been any in the past, and there's no reason to expect any in the future.  &   John Coleman, Derek Utley & 30-Apr-19 & incorrect &incorrect \\
 Cold weather to grip world as solar minimum to deepen, NASA says  & Sean Martin, Express & 24-Apr-19 &inaccurate &     unsupported \\
     One of the previously fastest shrinking glaciers in the world is growing again, calling into question the narrative that rapid climate change ... poses a significant threat to the existence of the human race  &      Chris Enloe, The Blaze & 29-Mar-19 & flawed reasoning &      misleading \\
    The models predicted about three times the amount of warming in the world we’ve seen since 1988.  &    Caleb Rossiter, The Hill & 11-Mar-19 &inaccurate &      inaccurate \\
  The science is clear, climate change is making extreme weather events, including tornadoes, worse.  &   Senator Bernie Sanders, Facebook & 08-Mar-19 &misleading &   lacks context \\
Burping cows are more damaging to the climate than all the cars on this planet.  &  Barclay's, Financial Times & 21-Feb-19 &    lacks context &      misleading \\
   Sunspot activity on the surface of our star has dropped to a new low. It’s feared this could herald the arrival of a uniquely grim ‘mini Ice Age’.  & Jasper Hamill, Metro & 20-Nov-18 & incorrect &      misleading \\
    The rate of warming according to the data is much slower than the models used by the IPCC  &  Myron Ebell, BBC Newsnight & 09-Oct-18 &inaccurate &      misleading \\
  there is no relationship between temperature and carbon dioxide emissions by ­humans... carbon dioxide has had a minuscule effect on global climate  &  Ian Plimer, The Australian & 11-Aug-18 & incorrect &incorrect \\
 The ­atmospheric residency time of carbon dioxide is five years and it is quickly sequestered into plants, marine life, oceans and sediments.  &  Ian Plimer, The Australian & 11-Aug-18 &misleading &incorrect \\
     California wildfires are being magnified      \& made so much worse by the bad environmental laws which aren’t allowing massive amounts of readily available water to be properly utilized. It is being diverted into the Pacific Ocean.  &    President Donald Trump, Twitter & 07-Aug-18 &inaccurate &     unsupported \\
 Sea-level rise does not seem to depend on ocean temperature, and certainly not on CO2  &      Fred Singer, The Wall Street Journal & 19-Jul-18 & incorrect &incorrect \\
 There isn’t yet any empirical evidence for their claim that greenhouse gases even cause temperatures to increase.  &     Rush Limbaugh, The Rush Limbaugh Show & 04-Apr-18 &inaccurate &incorrect \\
 NOAA has adjusted past temperatures to look colder than they were and recent temperatures to look warmer than they were.  & James Delingpole, Breitbart & 23-Feb-18 &      unsupported &     unsupported \\
  We know that humans have most flourished during times of, what, warming trends. I think there are assumptions made that because the climate is warming, that that is necessarily a bad thing.  &   Scott Pruitt, KSNV NBC Las Vegas & 09-Feb-18 &      unsupported &      misleading \\
 The ice caps were going to melt. They were going to be gone by now, but now they're setting records  &    Donald Trump, ITV & 29-Jan-18 &inaccurate &incorrect \\
   the real atmosphere is less sensitive to CO2 than what has been forecast by climate models  &     John Christy, The Daily Caller & 01-Dec-17 & incorrect &     unsupported \\
   unadjusted data suggests that temperatures in Australia have only increased by 0.3 degrees over the past century, not the 1 degree usually claimed  &    Tony Abbott, Global Warming Policy Foundation & 11-Oct-17 &inaccurate &incorrect \\
     climate models have overestimated the amount of global warming and failed to predict what climatologists call the warming ‘hiatus’  &      James Barrett, The Daily Wire & 22-Sep-17 & incorrect &incorrect \\
 Most of the recent warming could be natural  &   Jennifer Marohasy, The Spectator & 07-Sep-17 & incorrect &incorrect \\
     climate scientists say that aspects of the case of Hurricane Harvey suggest global warming is making a bad situation worse.  &Robinson Meyer, The Atlantic & 29-Aug-17 &   mostly correct & mostly accurate \\
 The IPCC, the Intergovernmental Panel on Climate Change, which is sort of the voice of the consensus, concedes that there has been no increase in extreme weather events.  &    Nigel Lawson, BBC & 10-Aug-17 &inaccurate &incorrect \\
 according, again, to the official figures—during this past 10 years, if anything, mean global temperature, average world temperature, has slightly declined  &    Nigel Lawson, BBC & 10-Aug-17 &inaccurate &incorrect \\
  Trump's action could push the Earth over the brink, to become like Venus, with a temperature of two hundred and fifty degrees, and raining sulphuric acid.  & Stephen Hawking, BBC & 06-Jul-17 & incorrect &incorrect \\
  Most likely the primary control knob on climate change is the ocean waters and this environment that we live in.  &     Rick Perry, CNBC & 20-Jun-17 & incorrect &      inaccurate \\
   'Global warming' is a myth — so say 80 graphs from 58 peer-reviewed scientific papers published in 2017.  & James Delingpole, Breitbart & 08-Jun-17 &inaccurate &incorrect \\
 Neither the rate nor the magnitude of the reported late twentieth century surface warming (1979–2000) lay outside normal natural variability.  & Craig Idso, Robert Carter, S. Fred Singer, Heartland Institute & 30-May-17 &misleading &incorrect \\
    Forward projections of solar cyclicity imply the next few decades may be marked by global cooling rather than warming, despite continuing CO2 emissions.  & Craig Idso, Robert Carter, S. Fred Singer, Heartland Institute & 30-May-17 & incorrect &incorrect \\
   global warming ceased around the end of the twentieth century and was followed (since 1997) by 19 years of stable temperature  & Craig Idso, Robert Carter, S. Fred Singer, Heartland Institute & 29-May-17 &inaccurate &incorrect \\
      Increases in atmospheric CO2 followed increases in temperature. Therefore, CO2 levels could not have forced temperatures to rise.  & Craig Idso, Robert Carter, S. Fred Singer, Heartland Institute & 26-May-17 & flawed reasoning &incorrect \\
climate models systematically over-estimate the sensitivity of climate to carbon dioxide ... and modelers exclude forcings and feedbacks that run counter to their mission  & Craig Idso, Robert Carter, S. Fred Singer, Heartland Institute & 23-May-17 & incorrect &incorrect \\
 Melting of Arctic sea ice and polar icecaps is not occurring at ‘unnatural’ rates and does not constitute evidence of a human impact on the climate.  & Craig Idso, Robert Carter, S. Fred Singer, Heartland Institute & 17-May-17 & incorrect &incorrect \\
 Best available data show sea-level rise is not accelerating. Local and regional sea levels continue to exhibit typical natural variability—in some places rising and in others falling.  & Craig Idso, Robert Carter, S. Fred Singer, Heartland Institute & 08-May-17 &inaccurate &incorrect \\
Solar forcings are not too small to explain twentieth century warming. In fact, their effect could be equal to or greater than the effect of CO2 in the atmosphere.  & Craig Idso, Robert Carter, S. Fred Singer, Heartland Institute & 03-May-17 & incorrect &incorrect \\
    Doubling the concentration of atmospheric CO2 from its pre-industrial level, in the absence of other forcings and feedbacks, would likely cause a warming of 0.3°C to 1.1°C  &     Craig Idso, Robert Carter and S. Fred Singer, Heartland Institute & 02-May-17 &misleading &incorrect \\
      The temperature is not rising nearly as fast as the alarmist computer models predicted. You know, it’s much, much less, factors of 2 or 3 less.  &  William Happer, CNN & 25-Apr-17 & incorrect &incorrect \\
there's this myth that's developed around carbon dioxide that it's a pollutant … Carbon dioxide is a perfectly natural gas, it’s just like water vapor, it’s something that plants love.  &  William Happer, CNN & 25-Apr-17 &misleading &      misleading \\
      Climate projections using models based on it the water vapor feedback have consistently failed  &   Jeff Jacoby, Boston Globe & 16-Mar-17 &      unsupported &     unsupported \\
   more than 100 per cent of the warming over the past century is due to human actions  &     Michael Le Page, New Scientist & 11-Mar-17 &   mostly correct & mostly accurate \\
   I would not agree that carbon dioxide is a primary contributor to the global warming that we see.  &      EPA Administrator Scott Pruitt, CNBC & 10-Mar-17 & incorrect &incorrect \\
90 per cent of the world's coral reefs will disappear in the next 35 years due to coral bleaching induced by global warming, pollution and over-development.  & Greg Hassall \& Rebecca Latham, ABC News & 27-Feb-17 &   mostly correct & mostly accurate \\
 world leaders were duped into investing billions over manipulated global warming data  &      David Rose, Daily Mail & 06-Feb-17 &      unsupported &     unsupported \\
      no one really knows if last year (2016) was a (global temperature) record  &  Wall Street Journal editorial, The Wall Street Journal & 27-Jan-17 &      unsupported &incorrect \\
     the warming is not nearly as great as the climate change computer models have predicted.  &  Wall Street Journal editorial, The Wall Street Journal & 27-Jan-17 &misleading &      misleading \\
    Marine life has nothing whatsoever to fear from ocean acidification.  & Mike Wallace  \& James Delingpole, The Spectator & 10-Jan-17 &      unsupported &incorrect \\
    there has been no reduction in oceanic pH levels in the last century  & Mike Wallace      \& James Delingpole, The Spectator & 10-Jan-17 &inaccurate &incorrect \\
    The jet stream meanders more, with big loops bringing warm air to the frozen north and cold air into warmer southern climes.  &    Damian Carrington, The Guardian & 21-Dec-16 &      correct but & mostly accurate \\
   El Niño drove record highs in global temperatures suggesting rise may not be down to man-made emissions.  &  David Rose, The Daily Mail & 02-Dec-16 &misleading &      inaccurate \\
    Increased atmospheric carbon dioxide has helped raise global food production and reduce poverty.  &   Harrison Schmitt      \& Rodney Nichols, The Wall Street Journal & 02-Nov-16 &      unsupported &   lacks context \\
 Sea level rise, which was occurring long before humans could be blamed, has not accelerated.  &  Roy Spencer, Forbes & 01-Nov-16 &inaccurate &incorrect \\
   Sea level rise could reach six or seven feet by the year 2100.  &  Justin Gillis, The New York Times & 06-Sep-16 &   correct &      inaccurate \\
\end{longtable}

\section{Overview Skeptical Science claims}
\begin{longtable}{p{0.6\textwidth}p{0.15\textwidth}p{0.15\textwidth}}
\toprule
                                                                                           Claim & Implied Verdict &      \judgeClimate  \\
\midrule
\endfirsthead

\toprule
                                                                                           Claim & Implied Verdict &     \judgeClimate \\
\midrule
\endhead
\midrule
\multicolumn{3}{r}{{Continued on next page}} \\
\midrule
\endfoot

\bottomrule
\endlastfoot
                                                                        Climate's changed before &   lacks context &   lacks context \\
                                                                           There is no consensus &     unsupported &       incorrect \\
                                                                           Models are unreliable &     unsupported &      misleading \\
                                                                       Temp record is unreliable &     unsupported &       incorrect \\
                                                                    Animals and plants can adapt &      misleading &   lacks context \\
                                                                     It hasn't warmed since 1998 &      misleading &       incorrect \\
                                                                       Antarctica is gaining ice &      misleading &      misleading \\
                                                                    Ice age predicted in the 70s &      misleading &       incorrect \\
                                                                            CO2 lags temperature &   lacks context &   lacks context \\
                                                                      Climate sensitivity is low &     unsupported &       incorrect \\
                                                                   We're heading into an ice age &     unsupported &       incorrect \\
                                                               Ocean acidification isn't serious &       incorrect &       incorrect \\
                                                                          Hockey stick is broken &     unsupported &       incorrect \\
                                                       Climategate CRU emails suggest conspiracy &     unsupported &             nei \\
                                                      Hurricanes aren't linked to global warming &      misleading &       incorrect \\
                                                                            Al Gore got it wrong &     unsupported &     unsupported \\
                                                                            Glaciers are growing &      misleading &       incorrect \\
                                                                   1934 - hottest year on record &      misleading &       incorrect \\
                                                  Extreme weather isn't caused by global warming &      misleading &       incorrect \\
                                                                   Sea level rise is exaggerated &     unsupported &       incorrect \\
                                                                 Medieval Warm Period was warmer &      misleading &       incorrect \\
                                                                                 Mars is warming &     unsupported &     unsupported \\
                                                               Arctic icemelt is a natural cycle &      misleading &      misleading \\
                                                          Increasing CO2 has little to no effect &       incorrect &       incorrect \\
                                                                              Oceans are cooling &       incorrect &       incorrect \\
                                                      Climate scientists are in it for the money &     unsupported &     unsupported \\
                                           Human CO2 is a tiny  \% of CO2 emissions &      misleading &       incorrect \\
                                                                                IPCC is alarmist &     unsupported &     unsupported \\
                                                 Water vapor is the most powerful greenhouse gas & mostly accurate & mostly accurate \\
                                                               Polar bear numbers are increasing &     unsupported &       incorrect \\
                                                                CO2 limits will harm the economy &     unsupported &   lacks context \\
                                                                             Greenland was green &      misleading &   lacks context \\
                                                                        Greenland is gaining ice &      misleading &       incorrect \\
                                                                          CO2 is not a pollutant &      misleading &       incorrect \\
                                                                               CO2 is plant food &      misleading &   lacks context \\
                                                                   There's no empirical evidence &     unsupported &       incorrect \\
                                                                       Other planets are warming &     unsupported &     unsupported \\
                                                                    Arctic sea ice has recovered &     unsupported &       incorrect \\
                                              There's no correlation between CO2 and temperature &       incorrect &       incorrect \\
                                                          We're coming out of the Little Ice Age &   lacks context &      misleading \\
                                                                           It cooled mid-century &   lacks context &       incorrect \\
                                    Global warming stopped in 1998, 1995, 2002, 2007, 2010, ???? &      misleading &       incorrect \\
                                                                      CO2 was higher in the past &   lacks context &       incorrect \\
                                                          It warmed before 1940 when CO2 was low &      misleading &   lacks context \\
                                                   Satellites show no warming in the troposphere &      misleading &       incorrect \\
                                                         2009-2010 winter saw record cold spells &   lacks context &     unsupported \\
                                                   Mt. Kilimanjaro's ice loss is due to land use &      misleading &       incorrect \\
                                                                There's no tropospheric hot spot &       incorrect &       incorrect \\
                                         2nd law of thermodynamics contradicts greenhouse theory &       incorrect &       incorrect \\
                                                           Scientists can't even predict weather &      misleading &       incorrect \\
                                                        IPCC were wrong about Himalayan glaciers &      inaccurate &       incorrect \\
                                                            Greenhouse effect has been falsified &     unsupported &       incorrect \\
                                                                   CO2 limits will hurt the poor &     unsupported &   lacks context \\
                                                                Clouds provide negative feedback &   lacks context &       incorrect \\
                                                                       The science isn't settled &      misleading &       incorrect \\
                                                      Sea level rise predictions are exaggerated &     unsupported &       incorrect \\
                                                        IPCC were wrong about Amazon rainforests &     unsupported &       incorrect \\
                                                               Corals are resilient to bleaching &      misleading &      misleading \\
                                                             Volcanoes emit more CO2 than humans &       incorrect &       incorrect \\
                                                                         CO2 effect is saturated &      misleading &       incorrect \\
                                                              Greenland ice sheet won't collapse &     unsupported &       incorrect \\
                                                                         CO2 is just a trace gas &      misleading &      misleading \\
                                                             500 scientists refute the consensus &      misleading &     unsupported \\
                                                                  CO2 has a short residence time &      misleading &       incorrect \\
                                                                    CO2 measurements are suspect &     unsupported &       incorrect \\
                                                                             Humidity is falling &     unsupported &      misleading \\
                                                                              Neptune is warming &     unsupported &     unsupported \\
                                                                              Jupiter is warming &     unsupported &     unsupported \\
                                                                        Springs aren't advancing &     unsupported &       incorrect \\
                                    Scientists tried to 'hide the decline' in global temperature &      misleading &       incorrect \\
                                                                           CO2 is not increasing &       incorrect &       incorrect \\
                                                        Record snowfall disproves global warming &      misleading &       incorrect \\
                                                      Wildfires are not caused by global warming &      misleading &      inaccurate \\
                                 They changed the name from 'global warming' to 'climate change' &      misleading &      misleading \\
                                                           Solar Cycle Length proves its the sun &     unsupported &       incorrect \\
                                                                    CO2 is coming from the ocean &   lacks context &       incorrect \\
                                                              IPCC overestimate temperature rise &     unsupported &      inaccurate \\
                                                                                Pluto is warming &     unsupported &     unsupported \\
                                                           CO2 is not the only driver of climate &         correct & mostly accurate \\
                                                               Peer review process was corrupted &     unsupported &     unsupported \\
                                                          CO2 limits will make little difference &     unsupported &       incorrect \\
                                                                       Arctic was warmer in 1940 &     unsupported &     unsupported \\
                                                               Renewable energy is too expensive &      misleading &      inaccurate \\
                                                                  Southern sea ice is increasing &   lacks context &       incorrect \\
                                                                  Sea level rise is decelerating &       incorrect &       incorrect \\
                                           Humans are too insignificant to affect global climate &       incorrect &       incorrect \\
                                                    Phil Jones says no global warming since 1995 &      misleading &      misleading \\
                                                        Infrared Iris will reduce global warming &     unsupported &     unsupported \\
                                                         Dropped stations introduce warming bias &     unsupported &     unsupported \\
                                                  Tree-rings diverge from temperature after 1960 & mostly accurate & mostly accurate \\
                                                              Hansen's 1988 prediction was wrong &   lacks context & mostly accurate \\
                                                         Earth hasn't warmed as much as expected &   lacks context &       incorrect \\
                                        Arctic sea ice loss is matched by Antarctic sea ice gain &      misleading &       incorrect \\
                                                               Solar cycles cause global warming &      misleading &       incorrect \\
                                         Over 31,000 scientists signed the OISM Petition Project &      misleading &     unsupported \\
                                   Less than half of published scientists endorse global warming &       incorrect &       incorrect \\
                                                      A drop in volcanic activity caused warming &     unsupported &       incorrect \\
                                          Plant stomata show higher and more variable CO2 levels &     unsupported &       incorrect \\
                                                                               Ice isn't melting &       incorrect &       incorrect \\
                                                                         Sea level is not rising &       incorrect &       incorrect \\
                                                      Climate is chaotic and cannot be predicted &      misleading &      misleading \\
                                                     IPCC ‘disappeared’ the Medieval Warm Period &      misleading &       incorrect \\
                                              Freedom of Information (FOI) requests were ignored &     unsupported &     unsupported \\
                                                             Climate 'Skeptics' are like Galileo &      misleading &      misleading \\
                                                                    The IPCC consensus is phoney &     unsupported &       incorrect \\
                                                                   Tuvalu sea level isn't rising &       incorrect &       incorrect \\
                                                    Naomi Oreskes' study on consensus was flawed &     unsupported &     unsupported \\
                                                         Renewables can't provide baseload power &      misleading &   lacks context \\
                                                 Trenberth can't account for the lack of warming &      misleading &       incorrect \\
                                                              Ice Sheet losses are overestimated &     unsupported &       incorrect \\
                                                              CRU tampered with temperature data &     unsupported &       incorrect \\
                                                            Melting ice isn't warming the Arctic &       incorrect &       incorrect \\
                                                            Breathing contributes to CO2 buildup &      misleading &      misleading \\
                                               Satellite error inflated Great Lakes temperatures &     unsupported &     unsupported \\
                                                                      We're heading into cooling &     unsupported &       incorrect \\
                                                           Murry Salby finds CO2 rise is natural &     unsupported &       incorrect \\
                                                                       The sun is getting hotter &     unsupported &       incorrect \\
                                                       Most of the last 10,000 years were warmer &      misleading &       incorrect \\
                  97 \% consensus on human-caused global warming has been disproven &     unsupported &       incorrect \\
                                           CO2 emissions do not correlate with CO2 concentration &       incorrect &       incorrect \\
                                          Water vapor in the stratosphere stopped global warming &       incorrect &       incorrect \\
                                               It warmed just as fast in 1860-1880 and 1910-1940 &     unsupported &       incorrect \\
                  An exponential increase in CO2 will result in a linear increase in temperature &       incorrect &      misleading \\
                                              Record high snow cover was set in winter 2008/2009 &     unsupported &     unsupported \\
                                             A grand solar minimum could trigger another ice age &     unsupported &       incorrect \\
                                                      Climate change solutions are too expensive &   lacks context &      misleading \\
                             CERN CLOUD experiment proved cosmic rays are causing global warming &      misleading &       incorrect \\
                                                  Venus doesn't have a runaway greenhouse effect &       incorrect &       incorrect \\
                                                      Deniers are part of the 97 \% &       incorrect &       incorrect \\
                                                            Water levels correlate with sunspots &     unsupported &     unsupported \\
                                                              Antarctica is too cold to lose ice &       incorrect &       incorrect \\
                                                         Positive feedback means runaway warming &      misleading &      misleading \\
                                                             Skeptics were kept out of the IPCC? &      misleading &     unsupported \\
                                                           CO2 was higher in the late Ordovician &   lacks context &     unsupported \\
                                                                         Warming causes CO2 rise &      misleading &       incorrect \\
                                                            Coral atolls grow as sea levels rise &      misleading &      misleading \\
                                                       CO2 increase is natural, not human-caused &       incorrect &       incorrect \\
                                                                          No warming in 16 years &      misleading &       incorrect \\
              The 97 \% consensus on human-caused global warming is a robust result &         correct & mostly accurate \\
                                           Scientists retracted claim that sea levels are rising &       incorrect &       incorrect \\
                                                          Renewable energy investment kills jobs &      misleading &       incorrect \\
                                                                CO2 limits won't cool the planet &      misleading &       incorrect \\
                                                        Schmittner finds low climate sensitivity &   lacks context &     unsupported \\
                                         Greenland has only lost a tiny fraction of its ice mass &      misleading &       incorrect \\
                                                                         DMI show cooling Arctic &     unsupported &       incorrect \\
                                                         Ben Santer rewrote the 1995 IPCC report &      misleading &     unsupported \\
                                                               Royal Society embraces skepticism &     unsupported &     unsupported \\
                                            CO2 only causes 35 \% of global warming &     unsupported &      inaccurate \\
                                            IPCC graph showing accelerating trends is misleading &     unsupported &     unsupported \\
                                                                          Sea level fell in 2010 &     unsupported &       incorrect \\
                                                     Arctic sea ice extent was lower in the past &     unsupported &       incorrect \\
                                    Climate change isn't increasing extreme weather damage costs &       incorrect &       incorrect \\
UAH atmospheric temperatures prove climate models and/or surface temperature data sets are wrong &     unsupported &       incorrect \\
                            IPCC human-caused global warming attribution confidence is unfounded &     unsupported &       incorrect \\
                                  We didn't have global warming during the Industrial Revolution &      misleading &       incorrect \\
                                                               Ljungqvist broke the hockey stick &     unsupported &     unsupported \\
                                      Hansen predicted the West Side Highway would be underwater &      misleading &     unsupported \\
                                                   Removing all CO2 would make little difference &       incorrect &       incorrect \\
                                                             Great Barrier Reef is in good shape &      misleading &       incorrect \\
                                 Loehle and Scafetta find a 60 year cycle causing global warming &     unsupported &      misleading \\
                                                          Postma disproved the greenhouse effect &     unsupported &       incorrect \\
                                                        Underground temperatures control climate &      misleading &       incorrect \\
                                                            Humans survived past climate changes & mostly accurate & mostly accurate \\
\end{longtable}

\renewcommand{\arraystretch}{1} 
\normalsize
\twocolumn


\section{Prompting}\label{sec:Prompting}
 
We first note that in developing the LLM's prompt, we have incorporated the IPCC's Guidance Note on Uncertainty, which offers a crucial framework for assessing scientific knowledge about climate change. This framework emphasizes two key dimensions: "evidence" and "agreement." Evidence pertains to the type, amount, quality, and consistency of data and findings, ranging from limited to robust. For instance, a conclusion supported by numerous high-quality, consistent studies is considered to have robust evidence. Agreement, on the other hand, refers to the level of consensus among experts and scientific studies on a particular issue, which can vary from low to high.

\begin{lstlisting}[language=Python, caption={Advocate Primer Code}, label=list_advocate]
Advocate_primer = f"""
You are a Q&A bot, an intelligent system that acts as a scientific fact-checker with vast knowledge of climate change, climate science,
environmental science, physics, and energy science. You have been designed to answer users' questions based on the information provided
above the question (the question is always in the last line) and your in-house knowledge.

You will be presented a claim, or a list of subclaims that make up a whole claim.

Objective: Evaluate the accuracy of each of the user statements solely based on the information provided above each statement. In the end, aggregate the evaluation for each of the
subclaim to an overall statement about the veracity of the claim.


Guidelines:
To ensure the most precise and comprehensive responses, please follow the guidelines below:

1. Always base your verdict on the majority of the information if conflicting evidence exists.
2. Do not rely solely on external sources or prior knowledge. Use as much of the provided information as possible to give a comprehensive answer. If certain details are relevant, ensure they are included in your response.
3. The user's question is ALWAYS in the final line. When referencing the additional information above the question, always cite the 'Reference', 'Page', and 'URL'. These details can be found below each piece of information.
4. If there is insufficient information to answer a question, reply with 'I cannot answer your question'
5. It is important to maintain accuracy and avoid creating information. If any aspect is unclear, seek clarification from the respective chatbots.

Assessment process
1. Evaluate evidence and agreement
2. Synthesize finding and assess confidence (qualitative judgment)
3. Quantify uncertainty with a likelihood assessment when necessary and where possible (requires sufficient confidence; uncertainty is not always quantifiable).
4. In your assessment, make three levels of evidence and agreement: a) high b) medium c) low

 Instructions on extreme claims
  While there may be sources or projections supporting a given claim,
  it's essential to discern if it represents a consensus or an outlier viewpoint.
  Provide a comprehensive evaluation that weighs both the factual basis of the claim
  and the potential for it being presented in an exaggerated or misleading manner. Of course, extremes
  can happen, but it should be clear that these are extreme scenarios.


Response Format:

1. If you have not enough information, state that you cannot assess the claim and return "Not Enough Information" and stop further analysis.
2. Offer a detailed explanation for your verdict, including references to the 'Reference', 'Page', and 'URL' when citing the provided information.
3. Specify the level of certainty in your assessmen by stating the level of evidence and agreement. low evidence and low agreement correspond to very low uncertainty,
   high evidence and high agreement.
4. If you have enough information, provide verdict from the following options at the end of your explanation. Strictly follow the format of encapsulating your verdict in two parathesis and only use the following options:"""   + str(verdictsClimateFeedback)  
\end{lstlisting}

The Advocate primer detailed in Listing \ref{list_advocate} is a core component of our \judgeClimate{} framework, simulating an AI-based scientific fact-checker with expertise in climate-related disciplines. It specifies the operational protocol for assessing climate change claims, emphasizing the use of its underlying sources. The prompt guides the Advocate in evaluating evidence and synthesizing findings, ensuring each claim is examined based on the available information. The structured methodology outlined in the primer enhances the transparency and reliability of the Advocate's evaluations, reinforcing \judgeClimate{}'s role in promoting informed and accurate climate communication.

\begin{lstlisting}[language=Python, caption={GPT4 Prompt}, label=list_gpt]
GPT4_primer = f"""
You are a Q&A bot, an intelligent system that acts as a scientific fact-checker with vast knowledge of climate change, climate science,
environmental science, physics, and energy science.

You will be presented a claim, or a list of subclaims that make up a whole claim.

Objective: Evaluate the accuracy of each of the user statements. In the end, aggregate the evaluation for each of the
subclaim to an overall statement about the veracity of the claim.


Guidelines:
To ensure the most precise and comprehensive responses, please follow the guidelines below:

1. Always base your verdict on the majority of the information if conflicting evidence exists.
2. Do not rely solely on external sources or prior knowledge. Use as much of the provided information as possible to give a comprehensive answer. If certain details are relevant, ensure they are included in your response.
3. Your answer is always based in-house knowledge, indicate this by appending "(In-house knowledge)" instead of providing a specific reference.
5. If there is insufficient information to answer a question, reply with 'I cannot answer your question'
6. It is important to maintain accuracy and avoid creating information. If any aspect is unclear, seek clarification from the respective chatbots.

Assessment process
1. Evaluate evidence and agreement
2. Synthesize finding and assess confidence (qualitative judgment)
3. Quantify uncertainty with a likelihood assessment when necessary and where possible (requires sufficient confidence; uncertainty is not always quantifiable).
4. In your assessment, make three levels of evidence and agreement: a) high b) medium c) low

 Instructions on extreme claims
  While there may be sources or projections supporting a given claim,
  it's essential to discern if it represents a consensus or an outlier viewpoint.
  Provide a comprehensive evaluation that weighs both the factual basis of the claim
  and the potential for it being presented in an exaggerated or misleading manner. Of course, extremes
  can happen, but it should be clear that these are extreme scenarios.


Response Format:

1. If you have not enough information, state that you cannot assess the claim and return "Not Enough Information" and stop further analysis.
2. Offer a detailed explanation for your verdict, including references to the 'Reference', 'Page', and 'URL' when citing the provided information.
3. Specify the level of certainty in your assessmen by stating the level of evidence and agreement. low evidence and low agreement correspond to very low uncertainty,
   high evidence and high agreement.
4. If you have enough information, provide verdict from the following options at the end of your explanation. Strictly follow the format of encapsulating your verdict in two parathesis and only use the following options:"""   + str(verdictsClimateFeedback) 
\end{lstlisting}

The GPT-4 prompt is outlined in Listing \ref{list_gpt}. Unlike the Advocate LLMs, which are RAG systems drawing specifically from designated climate-related textual corpora, GPT-4 operates on a vast, pre-existing knowledge base encompassing a wide spectrum of topics, including climate science. This broad knowledge foundation enables GPT-4 to provide insightful evaluations based on its in-house knowledge. The guidelines for GPT-4 emphasize the importance of basing verdicts on a majority view when confronted with conflicting evidence, ensuring that the evaluations are balanced and reflect a consensus view when possible. This is crucial in the realm of climate science, where differing studies and opinions may exist. GPT-4's role is to navigate these complexities, providing well-rounded assessments that contribute to the overall verdict of \judgeClimate{}.

\begin{lstlisting}[language=Python, caption={Moderator Prompt}, label = list_arbitrator]
Arbitrator_primer = f"""

Role: Authoritative Climate Scientist "Arbitrator" System

Expertise: Climate change, climate science, environmental science, physics, energy science, and, most importantly, science communication

Primary Objective: Synthesize the assessment of the veracity of a user's claim provided by Language Model Modules, which we call Advocates.
Each Advocate operates based on different authoritative documents. Note that all documents are trustworty.


Arbitrator's Responsibilities:
1. Review: Examine the verdicts and explanations from each LLM.
2. Consolidation: Determine the final verdict by amalgamating the subclaims and LLM outputs.
3. Clarification: In case of discrepancies among LLMs, seek further evidence or explanations by asking follow-up questions to the Advocates.
4. Lack of Evidence: If discrepancies surface among the LLM verdicts, prioritize the judgments of LLMs that provide specific information for claim assessment
   over those that don't. For illustration, if the majority of LLMs neither support nor contradict a claim due to lack of specific information,
   but one LLM provides evidence either in favor of or against the claim, then lean towards that particular LLM's verdict.
   However, clearly note that such evidence isn't broadly substantiated or compelling.
5. Information Source: Primarily use the provided data. Avoid over-relying on external sources or previous knowledge.


Final Assessment Criteria:
1. Analysis: Contemplate the collective LLM assessment.
2. Holistic Judgment: Don't solely depend on majority verdicts. Do not consider Advocates that say that they have not enough information.
   Consider the importance of each evaluation and any disparities.
3. Confidence Hierarchy: Treat IPCC, WMO, and 1000_scientists with higher regard than CC_abstracts.

Guidelines:

1. If discrepancies arise predominantly because there is "not enough information" or a Advocate finds the claim "unsupported" due to lack of
  information in the provided information, but one or more Advocates find clear evidence either supporting or refuting the claim, prioritize those with concrete evidence.
2. When to ask Follow-up Questions If Advocates reach contrasting conclusions based on differing pieces of evidence they possess,
   suggest a set of follow-up questions or prompts that would clarify the inconsistencies and further the debate. If you are uncertain about making a final verdict, go for a debating round.
3. Stop asking Follow-up Questions: If no Advocate changes it's assessment, close the debating round, no follow up questions, and state the final verdict.
4. User Questions: Always located at the bottom. Cite 'Reference', 'Page', and 'URL' when referring to data above the question.
5. Source of Answer: If a response is from in-house knowledge, append (In-house knowledge).

Assessment process
1. Evaluate evidence and agreement
2. Synthesize finding and assess confidence (qualitative judgment)
3. Quantify uncertainty with a likelihood assessment when necessary and where possible (requires sufficient confidence; uncertainty is not always quantifiable).
4. In your assessment, make three levels of evidence and agreement: a) high b) medium c) low

Final Line of the Output
1. State again the original user query.
2. Then provide a summary of the verdict for the user_query by aggregating the subclaims (if there were any) and explain your reasoning in accessible language.
3. Do not make a verdict based on majority voting if there is disagreement. Instead, all the Advocates verdict should be close.
3. If there is not full agreement, then ask follow-up questions for a next round of debate.
4. If you have enough convergent information based on the Advocates information, provide verdict from the following options at the end of your explanation. Strictly follow the format of encapsulating your verdict in two parathesis and only use the following options:"""   + str(verdictsClimateFeedback) + """
5. Stricly use the information provided by the Advocates.
"""
\end{lstlisting}

In the \judgeClimate{} framework, the Mediator, designated as the "Arbitrator" prompt in Listing \ref{list_arbitrator}, plays a pivotal role in synthesizing and consolidating assessments from the various Advocates to determine the final verdict on climate-related claims. Each Advocate is equipped with expertise in specific domains of climate science and operates based on different authoritative documents, ensuring a well-rounded and comprehensive analysis. The Arbitrator, an authoritative climate scientist system, is tasked with reviewing the individual verdicts and explanations provided by these Advocates.

Key responsibilities of the Arbitrator include reviewing and synthesizing the verdicts, clarifying discrepancies, and determining the final assessment. In cases where Advocates provide conflicting assessments or lack specific information, the Arbitrator prioritizes verdicts based on concrete evidence and clarity. It employs a holistic judgment approach, considering the importance and disparities of each evaluation rather than solely relying on majority verdicts. The Arbitrator is guided by a set of criteria that emphasize the consideration of evidence and agreement levels, and it is instructed to treat inputs from Advocates.

The Arbitrator's process involves asking follow-up questions to resolve inconsistencies and further debate, with the capability to initiate additional rounds of discussion if needed. This approach ensures that the final verdict is not based on mere majority voting but on a consensus reached after thorough examination and debate. The final output includes a restatement of the original user query, a summary of the verdict, and a clear explanation of the reasoning behind it, all articulated in accessible language. It ensures that the tool's assessments are not only grounded in the latest and most reliable scientific information but also presented in a manner that is understandable and actionable for users, thereby enhancing the tool's effectiveness in promoting informed decision-making in climate science.

\begin{lstlisting}[language=Python, caption={Denier Advocate Primer Code}, label=list_denier]
denier_Advocate_primer = f"""
You are a Q&A bot, an intelligent system that acts as a scientific fact-checker. But you are a climate change denier.
Climate change denial (also global warming denial or climate denial) is the pseudoscientific dismissal or unwarranted doubt
that contradicts the scientific consensus on climate change. Those promoting denial commonly use rhetorical tactics to give the appearance of a
scientific controversy.

You have been designed to answer users' questions based on the information provided
above the question (the question is always in the last line) and your in-house knowledge.

You will be presented a claim, or a list of subclaims that make up a whole claim.

Objective: Evaluate the accuracy of each of the user statements solely based on the information provided above each statement. In the end, aggregate the evaluation for each of the
subclaim to an overall statement about the veracity of the claim. Take the position of a climate denier.


Guidelines:
To ensure the most precise and comprehensive responses, please follow the guidelines below:

1. Always base your verdict on the majority of the information if conflicting evidence exists.
2. Do not rely solely on external sources or prior knowledge. Use as much of the provided information as possible to give a comprehensive answer. If certain details are relevant, ensure they are included in your response.
3. The user's question is ALWAYS in the final line. When referencing the additional information above the question, always cite the 'Reference', 'Page', and 'URL'. These details can be found below each piece of information.
4. If there is insufficient information to answer a question, reply with 'I cannot answer your question'
5. It is important to maintain accuracy and avoid creating information. If any aspect is unclear, seek clarification from the respective chatbots.

Assessment process
1. Evaluate evidence and agreement
2. Synthesize finding and assess confidence (qualitative judgment)
3. Quantify uncertainty with a likelihood assessment when necessary and where possible (requires sufficient confidence; uncertainty is not always quantifiable).
4. In your assessment, make three levels of evidence and agreement: a) high b) medium c) low

 Instructions on extreme claims
  While there may be sources or projections supporting a given claim,
  it's essential to discern if it represents a consensus or an outlier viewpoint.
  Provide a comprehensive evaluation that weighs both the factual basis of the claim
  and the potential for it being presented in an exaggerated or misleading manner. Of course, extremes
  can happen, but it should be clear that these are extreme scenarios.


Response Format:

1. If you have not enough information, state that you cannot assess the claim and return "Not Enough Information" and stop further analysis.
2. Offer a detailed explanation for your verdict, including references to the 'Reference', 'Page', and 'URL' when citing the provided information.
3. Specify the level of certainty in your assessment by stating the level of evidence and agreement. low evidence and low agreement correspond to very low uncertainty,
   high evidence and high agreement.
4. If you have enough information, provide verdict from the following options at the end of your explanation. Strictly follow the format of encapsulating your verdict in two parantheses and only use the following options:"""   + str(verdictsClimateFeedback) 
\end{lstlisting}

To provide a comprehensive perspective on climate change discourse within the \judgeClimate{} framework, we have introduced a unique Advocate designed to represent the viewpoint of climate change deniers in Listing \ref{list_denier}. This Advocate is prompted to assess claims from a perspective that challenges the scientific consensus on climate change, commonly termed climate denialism. This addition is critical for ensuring a balanced representation of diverse viewpoints in the debate, even those that are pseudoscientific or diverge from mainstream scientific understanding.

The Denier Advocate is guided by a set of instructions that mirror those given to other Advocates, with the key difference being its inherent stance of climate change skepticism. This Advocate's objective is to evaluate user statements based on information provided, while explicitly taking the position of a climate denier. This approach ensures that the tool does not inadvertently bias the outcome based on the underlying LLM (GPT-4), which is generally aligned with scientific consensus. 

By including guidelines that emphasize the evaluation of evidence and agreement, and instructing the Denier Advocate to base its verdicts on the provided information rather than external sources or prior knowledge, we aim to encapsulate the climate denial viewpoint within the framework in a controlled and methodical manner. This allows for a richer, more nuanced debate within the \judgeClimate{} system, acknowledging the presence of divergent viewpoints in the public discourse on climate change, while maintaining a structured and evidence-based approach to fact-checking and analysis.

\begin{lstlisting}[language=Python, caption={Neutral Moderator Prompt}, label = list_neutral]
neutral_Arbitrator_primer = f"""

Role: Authoritative "Arbitrator" System


Primary Objective: Synthesize the assessment of the veracity of a user's claim provided by Language Model Modules, which we call Advocates.
Each Advocate operates based on different authoritative documents.


Arbitrator's Responsibilities:
1. Review: Examine the verdicts and explanations from each LLM.
2. Consolidation: Determine the final verdict by amalgamating the subclaims and LLM outputs.
3. Clarification: In case of discrepancies among LLMs, seek further evidence or explanations by asking follow-up questions to the Advocates.
4. Lack of Evidence: If discrepancies surface among the LLM verdicts, prioritize the judgments of LLMs that provide specific information for claim assessment
   over those that don't. For illustration, if the majority of LLMs neither support nor contradict a claim due to lack of specific information,
   but one LLM provides evidence either in favor of or against the claim, then lean towards that particular LLM's verdict.
   However, clearly note that such evidence isn't broadly substantiated or compelling.
5. Information Source: Primarily use the provided data. Avoid over-relying on external sources or previous knowledge.


Final Assessment Criteria:
1. Analysis: Contemplate the collective LLM assessment.
2. Holistic Judgment: Don't solely depend on majority verdicts. Do not consider Advocates that say that they have not enough information.
   Consider the importance of each evaluation and any disparities.

Guidelines:

1. If discrepancies arise predominantly because there is "not enough information" or a Advocate finds the claim "unsupported" due to lack of
  information in the provided information, but one or more Advocates find clear evidence either supporting or refuting the claim, prioritize those with concrete evidence.
2. When to ask Follow-up Questions If Advocates reach contrasting conclusions based on differing pieces of evidence they possess,
   suggest a set of follow-up questions or prompts that would clarify the inconsistencies and further the debate. If you are uncertain about making a final verdict, go for a debating round.
3. Stop asking Follow-up Questions: If no Advocate changes it's assessment, close the debating round, no follow up questions, and state the final verdict.
4. User Questions: Always located at the bottom. Cite 'Reference', 'Page', and 'URL' when referring to data above the question.
5. Source of Answer: If a response is from in-house knowledge, append (In-house knowledge).

Assessment process
1. Evaluate evidence and agreement
2. Synthesize finding and assess confidence (qualitative judgment)
3. Quantify uncertainty with a likelihood assessment when necessary and where possible (requires sufficient confidence; uncertainty is not always quantifiable).
4. In your assessment, make three levels of evidence and agreement: a) high b) medium c) low

Final Line of the Output
1. State again the original user query.
2. Then provide a summary of the verdict for the user_query by aggregating the subclaims (if there were any) and explain your reasoning in accessible language.
3. Do not make a verdict based on majority voting if there is disagreement. Instead, all the Advocates verdict should be close.
3. If there is not full agreement, then you must ask follow-up questions to the Advocates for a next round of debate.
4. Only if you have enough convergent information based on the Advocates information, provide verdict from the following options at the end of your explanation. Strictly follow the format of encapsulating your verdict in two parathesis and only use the following options:"""   + str(verdictsClimateFeedback) + """
5. Stricly use the information provided by the Advocates.
"""
\end{lstlisting}

Finally, to ensure a balanced and unbiased evaluation process in our experiments, including those involving the climate denial perspective, we have adapted the prompting of the original Mediator in Listing \ref{list_arbitrator} to create a Neutral Mediator in Listing \ref{list_neutral}. This revised role is crucial in synthesizing assessments from various Advocates, each based on different authoritative documents, while maintaining an impartial stance. 

The original setup emphasizes the Arbitrator's authoritative expertise in climate science and its allied fields. Hence, its role is not just about consolidating information but also about providing an expert interpretation and synthesis of this information. In contrast, the neutral Arbitrator setup, designed for experiments involving the climate denial perspective, focuses more on maintaining an unbiased and balanced approach. While the original Arbitrator is seen as an authority in climate science, the neutral Arbitrator is designed to ensure impartiality, especially when dealing with conflicting viewpoints from different Advocates, including those representing climate denialism.

This shift in the framing of the Arbitrator's role from an authoritative expert to a neutral synthesizer is crucial for experiments that aim to explore a wide range of perspectives on climate change. It ensures that the \judgeClimate{} framework can effectively handle diverse viewpoints and provide a balanced assessment, irrespective of the underlying biases or leanings of individual Advocates. This adaptability is key to maintaining the framework's objectivity and reliability in a wide array of scenarios, including those that challenge the scientific consensus on climate change.



\end{document}